\pdfoutput=1
\documentclass[11pt]{article}

\usepackage[final]{acl}

\usepackage{inconsolata}

\usepackage{hyperref}
\usepackage{url}
\usepackage{color,colortbl}
\usepackage{adjustbox}
\usepackage{graphicx}
\usepackage{tikz}
\usepackage{caption}
\usepackage{xcolor}

\definecolor{darkgreen}{rgb}{0,0.5,0}
\definecolor{azureblue}{rgb}{0,0.5,1}
\definecolor{darkgreen}{rgb}{1,0,0}
\definecolor{color1}{HTML}{006EB8}
\definecolor{color2}{HTML}{009B55}
\definecolor{color3}{HTML}{00A99A}
\definecolor{color4}{HTML}{3C8031}
\definecolor{color5}{HTML}{006795}
\definecolor{color6}{HTML}{00AEB3}
\definecolor{mygray}{gray}{0.93}
\definecolor{mygreen}{HTML}{3FBC9D}
\definecolor{arsenic}{rgb}{0.23, 0.27, 0.29}
\usepackage[utf8]{inputenc}
\usepackage[T1]{fontenc}
\usepackage{hyperref}
\usepackage{url} 
\usepackage{booktabs}
\usepackage{amsfonts}
\usepackage{nicefrac}
\usepackage{microtype}
\usepackage{wrapfig}
\usepackage[nointegrals]{wasysym}
\usepackage{lipsum}  
\usepackage{times}
\usepackage{tikz}
\usepackage{latexsym}
\usepackage{microtype}
\usepackage{graphicx}
\usepackage{placeins}
\usepackage{subcaption}
\usepackage{bbm}
\usepackage{multirow}
\usepackage{enumitem}
\usepackage{tikz}
\usepackage{amsmath}
\usepackage{amssymb}
\usepackage{mathtools}
\usepackage{amsthm}
\usepackage{nccmath}
\usepackage{algorithm}
\usepackage{caption}

\usepackage{listings}
\usepackage[algo2e, ruled]{algorithm2e}
\SetKwComment{Comment}{$\triangleright$\ }{}
\SetKwProg{Init}{Initialize}{}{}
\definecolor{color_green}{HTML}{009B55}
\usepackage{multirow}
\usepackage{pifont}
\usepackage{graphicx} % Required for inserting images
\usepackage{xspace}
\newcommand{\ourapproach}{\texttt{NPS}\xspace}
\newcommand{\pub}[1]{{\color{gray}{\tiny{[{#1}]\!}}}}

\newcommand{\thickhline}{\noalign{\hrule height 1.pt}}

\definecolor{mygray}{gray}{.9}
\definecolor{ggray}{RGB}{127,127,127}
\definecolor{reda}{RGB}{192,0,0}
\definecolor{redb}{RGB}{217,148,143}
\definecolor{myyellow}{RGB}{190,144,0}
\definecolor{mygreen}{RGB}{80,100,40}
\definecolor{myblue}{RGB}{30,90,100}
\definecolor{tabhighlight}{HTML}{e5e5e5}

\usepackage{fontawesome5}

\usepackage{seqsplit}

\title{Neural Parameter Search for Slimmer Fine-Tuned Models \\
and Better Transfer}

\author{
Guodong Du$^{1}$ \quad
        Zitao Fang$^{2}$ \quad
	Jing Li$^1$\textsuperscript{\faEnvelope} \quad
	Junlin Li$^{1}$ \quad 
	\textbf{Runhua Jiang}$^{2}$ \quad \\
	\textbf{Shuyang Yu}$^{2}$ \quad
	\textbf{Yifei Guo}$^{2}$ \quad
        \textbf{Yangneng Chen}$^{1}$ \quad 
        \textbf{Sim Kuan Goh}$^{2}$ \quad \\
        \textbf{Ho-Kin Tang}$^{1}$ \quad   
        \textbf{Daojing He}$^{1}$ \quad
        \textbf{Honghai Liu}$^{1}$ \quad
        \textbf{Min Zhang}$^{1}$ \quad \\
   $^{1}$Harbin Institute of Technology, Shenzhen, China\\
 	$^{2}$Xiamen University Malaysia \\
    \texttt{duguodong7@gmail.com} \quad \texttt{jingli.phd@hotmail.com} \quad \\ 
}

\begin{document}
\maketitle
\begin{abstract}
\label{sec:abstract}
Foundation models and their checkpoints have significantly advanced deep learning, boosting performance across various applications. 
However, fine-tuned models often struggle outside their specific domains and exhibit considerable redundancy. 
Recent studies suggest that combining a pruned fine-tuned model with the original pre-trained model can mitigate forgetting, reduce interference when merging model parameters across tasks, and improve compression efficiency.
In this context, developing an effective pruning strategy for fine-tuned models is crucial. 
Leveraging the advantages of the task vector mechanism, we preprocess fine-tuned models by calculating the differences between them and the original model. Recognizing that different task vector subspaces contribute variably to model performance, we introduce a novel method called \textbf{N}eural \textbf{P}arameter \textbf{S}earch (\textbf{\ourapproach}) for slimming down fine-tuned models. This method enhances pruning efficiency by searching through neural parameters of task vectors within low-rank subspaces.
Our method has three key applications: enhancing knowledge transfer through pairwise model interpolation, facilitating effective knowledge fusion via model merging, and enabling the deployment of compressed models that retain near-original performance while significantly reducing storage costs.
Extensive experiments across vision, NLP, and multi-modal benchmarks demonstrate the effectiveness and robustness of our approach, resulting in substantial performance gains. The code is publicly available at: \url{https://github.com/duguodong7/NPS-Pruning}.
\let\thefootnote\relax\footnotetext{\faEnvelope~Corresponding author.}	
\end{abstract}
\section{Introduction}
\label{sec:introduction}
In recent years, with the release of foundational models and the proliferation of associated checkpoints, the field of machine learning has undergone a paradigm shift. 
This shift has significantly enhanced the performance of downstream applications.
While fine-tuning pre-trained models~\cite{wortsman2022model,choshen2022fusing,liu2022tfew} has become common practice, these models often struggle with generalization and perform poorly outside their specific domains. Consequently, improving knowledge transfer from pre-trained to fine-tuned models has become a recent research focus~\cite{devlin2018bert}.
Consequently, recent research has increasingly focused on improving knowledge transfer, fusion, and compression by leveraging the parameters of the initial pre-trained model. Model Tailor~\cite{zhumodel} prunes fine-tuned models and combines them with the original model to reduce catastrophic forgetting. 
% Task Arithmetic~\cite{taskarithmetic} creates task vectors from the differences between fine-tuned and pre-trained models, which are then merged~\cite{du2024knowledge,jin2023regmean,singh2020model,wan2024knowledge,li2023deep} to improve knowledge fusion and multi-tasking. 
Additionally, TALL-masks~\cite{talls} compresses checkpoints by localizing task information within task vectors.
\begin{figure}[t]
    \centering
    \includegraphics[width=0.473\textwidth]{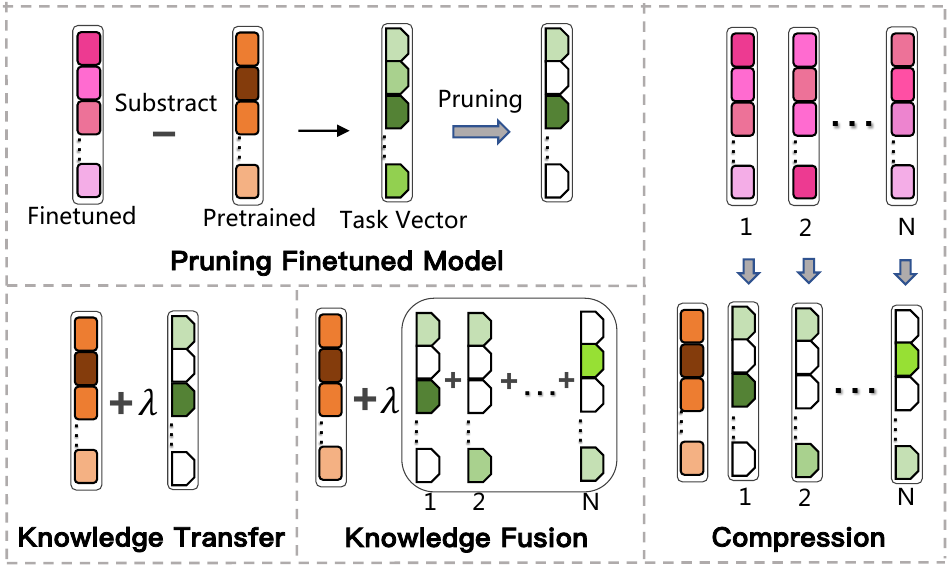}
    \caption{Knowledge transfer, fusion, and compression are enhanced with the assistance of pre-trained model parameters. The fine-tuned model is effectively represented as a combination of the pre-trained model and pruned task vectors, leading to knowledge retention.}
    \label{fig:overview}
    \vspace{-5mm}
\end{figure}
All these research efforts on knowledge transfer with available pre-trained parameters depend on a crucial preprocessing step: pruning the fine-tuned model task vectors~\cite{taskarithmetic}, as shown in Figure~\ref{fig:overview}. 

\begin{figure}[t]
    \centering
    \includegraphics[width=0.47\textwidth]{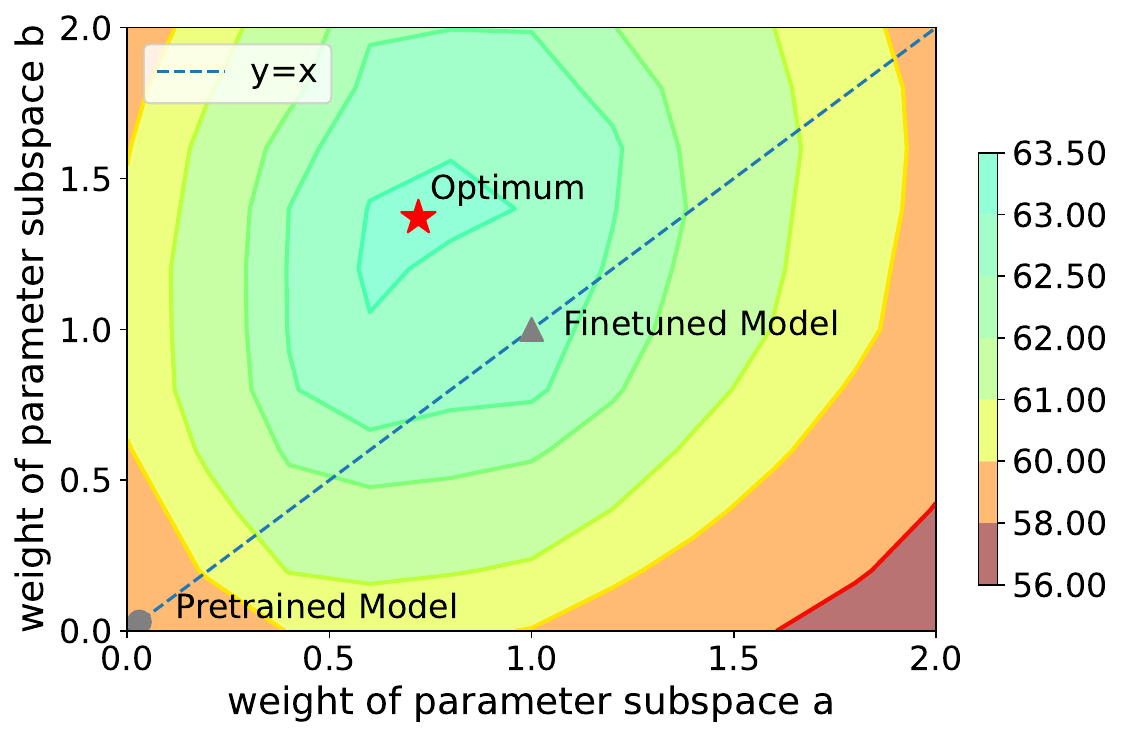}
    \caption{Performance of ViT-B/32 models on a specific task (SUN397 dataset). Different subspaces of neural parameters within the task vector contribute differently to the performance of the fine-tuned model.}
    \label{fig:motivation}
    \vspace{-2mm}
\end{figure}

Fine-tuned models often exhibit significant redundancy in parameter modifications compared to pre-trained models. Pruning these models can enhance the efficiency of knowledge representation.
Pruning fine-tuned model sets offers three main advantages: First, it reduces conflicts between fine-tuned models and the pre-trained model during knowledge transfer, thereby enhancing resilience to catastrophic forgetting. Second, it minimizes interference among fine-tuned models during fusion, improving multi-task generalization capabilities. Finally, pruning finetuned models can reduce storage costs while maintaining multi-task performance.
% Knowledge Transfer, Fusion and Compression with pretrained model available.
However, despite extensive research on model pruning in the context of compression~\cite{liang2021pruning,yu2023language,xia2022structured}, there is a relative scarcity of studies focused specifically on pruning fine-tuned models.
To address this gap, we propose a novel method called \textbf{N}eural \textbf{P}arameter \textbf{S}earch (\textbf{\ourapproach}) and design an adapted approach to apply pruned fine-tuned models in scenarios such as knowledge transfer, fusion, and compression. 
Specifically, we leverage the advantages of the task vector mechanism and preprocess fine-tuned models by calculating the difference between them and the original model. 
Recognizing that different task vector subspaces contribute variably to model performance, as shown in Figure~\ref{fig:motivation}, we search through the neural parameters within low-rank subspaces of task vectors. 
We partition the fine-tuned parameters into a set number of subspaces based on their magnitude, use evolutionary algorithms to assign new weights to different subspaces, and update the weights based on the model's performance on calibration datasets. 
This process avoids the need for gradient calculations, offering lightweight and efficient advantages.

We validated the effectiveness of our method across three key applications: knowledge transfer, model fusion, and compression. First, we performed interpolation between NPS-pruned models and the pre-trained model to mitigate forgetting, demonstrating superior performance on the multi-modal benchmark with LLaVa~\cite{zhumodel} model compared to previous methods.  
Second, we showed that weight averaging of multiple NPS-compressed fine-tuned models enables effective model fusion. Our approach was evaluated on NLP and vision tasks using models like T5~\cite{raffel2020exploring}, ViT~\cite{dosovitskiy2020image}, and LLaMa2~\cite{touvron2023llama}, as well as for fusing multiple PEFT adapters. Notably, it achieved a 4.3\% performance gain with T5-base.  
Finally, for deployment, our method allowed compressed models to retain near-original fine-tuned performance while significantly reducing storage costs. Extensive experiments demonstrated a 40\% improvement in compression efficiency on vision tasks.

% We validated the effectiveness of our method in three different application scenarios. %可以加上three different application scenarios: knowledge transfer, model fusion, and compression.
% First, we performed interpolation between the pruned models obtained through NPS and the pre-trained models to reduce the forgetting of the pre-trained models. 
% We tested the performance of the LLaVa model on the benchmark of multiple large language models and achieved performance that exceeded previous methods. 
% Additionally, we demonstrated that weight averaging of multiple NPS-compressed fine-tuned models can achieve model fusion. 
% We evaluated our approach across a range of NLP and vision tasks using various models, such as T5~\cite{raffel2020exploring}, ViT~\cite{dosovitskiy2020image}, and Llama2\cite{touvron2023llama}. 
% We also assessed its ability to fuse multiple PEFT adapters\cite{liu2022few, hu2021lora}. 
% All experiments showed significant improvements over previous state-of-the-art methods, notably achieving a 4.3\% performance increase with the T5-base model. 
% Finally, for deployment, specifying different compressed models allowed us to maintain almost the original fine-tuned performance while significantly reducing storage requirements. 
% We conducted extensive experimental testing and achieved notable improvements in storage efficiency, particularly with a 40\% increase in compression efficiency in experiments across 8 vision tasks. 
\vspace{2mm}
Our \textbf{contributions} can be summarized in the following four points:
\begin{itemize}[noitemsep,nolistsep]
    \item We reveal the importance of pruning fine-tuned models and highlight the limitations of previous methods.
    \item We propose \textbf{N}eural \textbf{P}arameter \textbf{S}earch (\textbf{\ourapproach}) for slimming down fine-tuned models.
    \item Based on the pruned fine-tuned models, we provide a simple and versatile method suitable for multi-task model fusion, compression, and robust knowledge transfer.
    \item Experimental results shown that our method significantly improves performance in various knowledge transfer scenarios.
\end{itemize}
\vspace{2mm}
\section{Related Work}
\label{sec:related Work}
\subsection{Knowledge Transfer, Fusion and Compression}
In the realms of knowledge transfer, model fusion \cite{jiang2024cade, fang2025disentangling}, and compression, foundational studies have driven significant progress.
~\cite{wortsman2022model} enhanced zero-shot learning by fine-tuning pre-trained models with minimal data, while~\cite{houlsby2019parameter} improved resource efficiency through parameter-efficient transfer learning.~\cite{chen2020simple} advanced model compression and fusion using contrastive learning in unsupervised settings, collectively marking major strides in model efficiency and robustness.

Recent years have seen the emergence of innovative methods for enhancing performance and efficiency across tasks when both pre-trained and fine-tuned models are available. 
Fisher-weighted averaging~\cite{matena2022merging} uses an information-theoretic approach to assess parameter importance, while RegMean~\cite{jin2022dataless} offers a closed-form solution for merging parameters through local linear regression. 
Task Arithmetic~\cite{ilharco2022editing}, PEM~\cite{zhang2023composing}, and TIES-Merging~\cite{ties} enhance model fusion through parameter composition, thereby improving model adaptability. Model Evolver~\cite{du2024knowledge, du2023end, du2024impacts} dynamically evolves model parameters, while Model Tailor~\cite{zhumodel} mitigates catastrophic forgetting in multimodal tasks through model patching, decoration, and post-training. Tall-masks~\cite{talls} offers efficient masking for model compression, and MATS~\cite{tam2024merging} employs a conjugate gradient method to match task parameter subspaces.

% In conclusion, our research builds on the approach of leveraging pre-trained models, as this strategy offers superior transfer performance and efficiency at a lower cost.
    \begin{figure*}[ht]
    \centering
    \includegraphics[width=\textwidth]{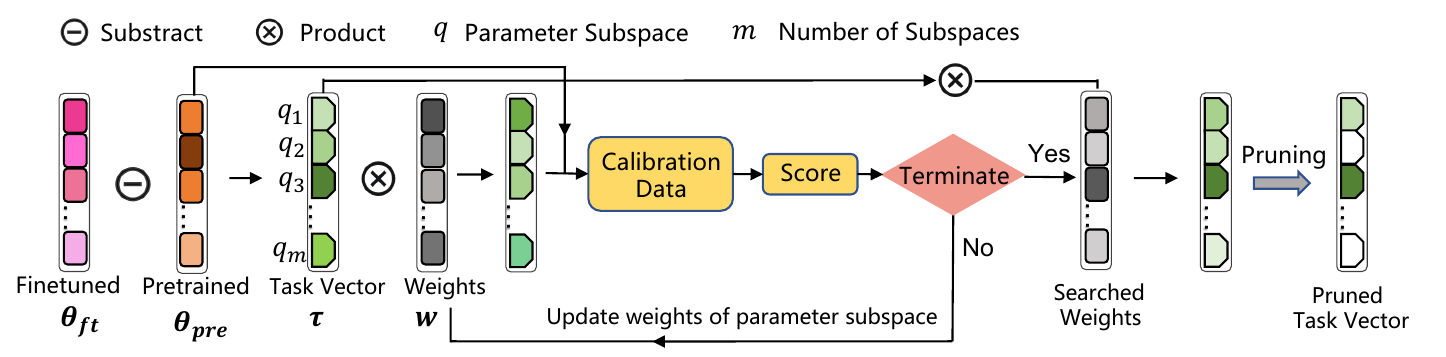}
    \caption{The framework of Neural Parameter Search enhances the efficiency of pruning fine-tuned models. This is achieved by searching and reweighting the neural parameters of task vectors within low-rank subspaces.}
    \label{fig:framework}
    \end{figure*}
In conclusion, our research focuses on leveraging pre-trained model parameters, as this approach provides better transfer performance and greater efficiency at a lower cost.

\subsection{Model Pruning}
% \subsection{Pruning Finetuned Models}
Model pruning can be broadly classified into two main approaches. The first approach encompasses traditional model pruning techniques. This includes structured pruning methods such as SliceGPT~\cite{slicegpt} and LLMpruner~\cite{llmpruner}, as well as unstructured pruning techniques like SparseGPT~\cite{frantar-sparsegpt}, Wanda~\cite{sun2023wanda}, GRAIN~\cite{yang-etal-2023-gradient}, GBLM-Pruner~\cite{das2023size}, and OWL~\cite{yin2023outlier}.

The second approach focuses on pruning fine-tuned models given a pretrained model. For instance, Model Grafting~\cite{panigrahi2023task} creates a mask to identify the most critical parameters for a specific task by optimizing the target task loss. TIES~\cite{ties} addresses interference issues that arise after magnitude pruning. DARE~\cite{yu2023language} aligns task vector parameters with the expected model output by randomly selecting and rescales them. Model Tailor~\cite{zhumodel} produces a sparse mask based on salience and sensitivity scores, while Talls Mask~\cite{talls} combines the merged model with an additional mask to localize task information, effectively reducing storage costs.

In this paper, we propose a novel pruning approach that is simple, efficient, and robust by searching for weight coefficients within neural parameter subspaces.

\section{Methodology}
\label{sec:method}
\subsection{Problem Setting}
 Here, we consider knowledge transfer, fusion and compression of a set of tasks \(\{T_1, \hdots, T_n\}\) and various pre-trained models like ViT~\cite{dosovitskiy2021an}, T5~\cite{raffel2020exploring}, or Llama2~\cite{touvron2023llama}. To begin, each pre-trained model is optimized on task-specific data, which can be performed either by fine-tuning the entire model or by using a parameter-efficient fine-tuning (PEFT) method~\cite{liu2022few, hu2021lora}. During this process, the trainable parameters \(\theta\) were initialized with \(\theta_\textrm{pre}\) (the pre-trained state) and subsequently updated to \(\theta_\textrm{ft}\) (the fine-tuned state).

Recent research introduced the concept of task vectors~\cite{ilharco2022editing}, which has been applied in various knowledge transfer, fusion, and compression tasks. 
For a specific task \( T \), the task vector \( \tau \in \mathbb{R}^d \) is defined as the difference between the fine-tuned weights \( \theta_i \) and the pre-trained weights \( \theta_{\text{pre}} \), i.e., \( \tau = \theta - \theta_{\text{pre}} \). 
This captures the changes during the fine-tuning phase for each task-specific model. 
Building on this idea, a pruned fine-tuned model $\hat{\theta_\textrm{ft}}$ can be obtained by first deriving the pruned task vector $\hat{\tau}$, as defined in the equation below:
\begin{equation}
\hat{\theta_\textrm{ft}} = \theta_{\text{pre}} + \hat{\tau}
\end{equation}

\subsection{Neural Parameter Search for Pruning}
Given that different parameter subspaces of task vectors contribute variably to fine-tuning performance, we first decomposed the task vector \(\tau\) into \(M\) independent parameter subspaces \(q_m\) by ranking the parameters based on their magnitude and then dividing them according to these ranks, summarized as \(\tau = \sum_{m=1}^{M} q_m\). 
Next, to enable more effective pruning, we reallocated weights for each subspace to obtain a new task vector:
\begin{equation}
\tau = \sum_{m=1}^{M}w_m*q_m
\end{equation}
while $*$ denotes scalar multiplication of a vector element-wise.

Initially, all weight coefficients were initialized to 1, after which we used an evolutionary algorithm to search for a more optimal set of weight coefficients. 
The optimization process aims to find the best set \(\{w_1, \hdots, w_m\}\), seeking optimal validation accuracy, and ultimately maximizing performance on calibration data with the adjusted fine-tuned model, as shown in Figure~\ref{fig:framework}.

In most of our experiments, we employed Covariance Matrix Adaptive Evolution Strategies (CMA-ES)~\cite{hansen1996adapting}, a probabilistic, population-based optimization algorithm. 
CMA-ES dynamically adjusts the search distribution through a covariance matrix, updating the mean and covariance at each iteration to effectively exploit the structure of the search space for obtaining optimal candidate solutions. 
When the evolutionary algorithm has approximately converged, we combined the optimized weight coefficients with the task vector and the pre-trained model to obtain an adjusted model:
\begin{equation}
\theta_\textrm{ft} = \theta_{\text{pre}} + \sum_{m=1}^{M}w_m*q_m
\end{equation}
Finally, we pruned the fine-tuned model based on the magnitude of its adjusted parameters after the search. We define the sparsity ratio as \( r \), where \( 0 < r \leq 1 \), and compute a mask \( m \) to select the most important neural parameters. This mask is derived using the following equation: 
\begin{equation}
m_{d} = \begin{cases} 
1, & \text{if } {\tau}_{d} \geq \text{sorted}({\tau})[r \times d] \\
0, & \text{otherwise}
\end{cases}
\end{equation}
The final pruned fine-tuned model is then given by: 
\begin{equation} \hat{\theta_\textrm{ft}} = \theta_{\text{pre}} + m \odot \tau
\end{equation}
while $\odot$ represents the Hadamard product.

\begin{figure}[t]
	\centering
\includegraphics[width=0.45\textwidth]{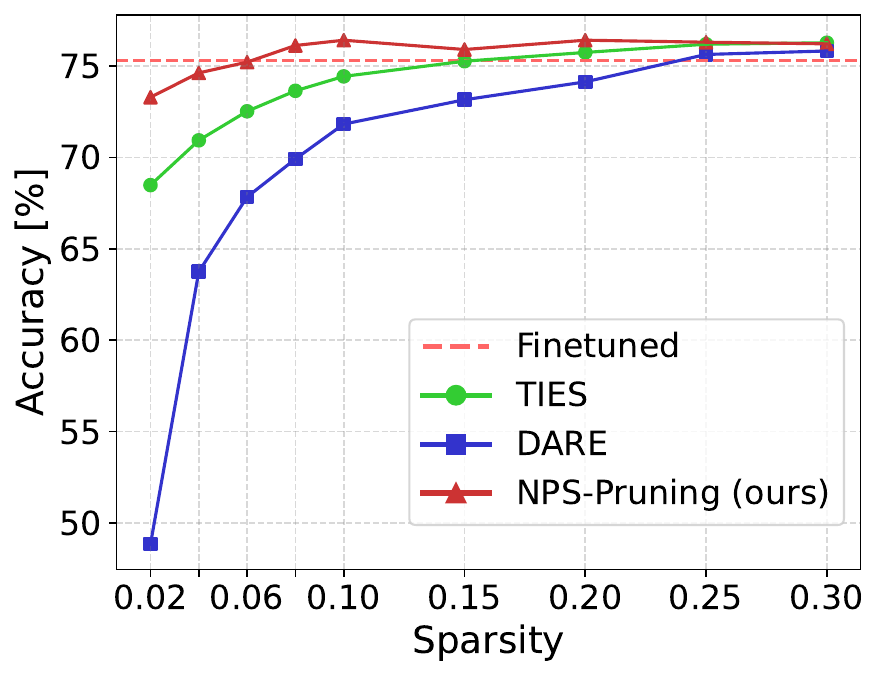}
	\caption{Performance variations of different methods with changes in sparsity ratio. Our \ourapproach method exhibits higher tolerance to varying levels of sparsity.}
	\label{fig:method}
    \vspace{-3mm}
    \end{figure}

This final model can subsequently be applied to scenarios such as knowledge transfer, fusion, and compression.
% $\tau = \sum_{m=1}^{M}q_m$
% The final pruned task vector is defined as \(\hat{\tau} = m \odot \tau\). 
To evaluate the pruning efficiency of the \ourapproach method, we applied it to a pre-trained vision model, ViT-B/32, which was fine-tuned on various tasks. We then assessed the results of different pruning methods on the respective benchmarks for each task. The reported results are the average performance across eight fine-tuned models under varying levels of pruning sparsity, as illustrated in Figure~\ref{fig:method}. In comparison with baseline methods like TIES~\cite{ties} and DARE~\cite{yu2023language}, our findings indicated that when the pruning sparsity ratio exceeds 0.2, most methods maintain performance comparable to the fine-tuned models. However, as the sparsity ratio drops below 0.2, accuracy tends to decline rapidly. Notably, our \ourapproach method demonstrates greater tolerance to lower sparsity ratios, preserving the original model's accuracy even at a sparsity ratio of 0.04.
\subsection{Applications}
\label{subsec:application}
Building on the significant improvement in pruning efficiency for fine-tuned models, we present three application scenarios for our proposed \ourapproach method in the context of knowledge transfer, model fusion, and compression when pre-trained models and task-specific data for fine-tuning are available.
\paragraph{Knowledge Transfer.}
Fine-tuning language models on new, unseen data often leads to a decline in performance on the original tasks. Moreover, previous research~\cite{zhumodel} indicates that fine-tuned models have low knowledge representation efficiency, containing a large number of redundant parameters that offer little benefit for new tasks. Removing these redundant parameters can minimize interference when integrating with the pre-trained model.
Therefore, combining a pruned fine-tuned model with the original pre-trained model can enhance its resistance to catastrophic forgetting during knowledge transfer. We propose applying \ourapproach to the parameters of the task vector before integrating them into the pre-trained model, as shown below:
\begin{equation} 
\hat{\theta}_{\textrm{ft}} = \theta_{\text{pre}} + \lambda \cdot m \odot \tau
\end{equation}
Here, \(\lambda\) is a hyperparameter used to rescale the neural parameters within the pruned task vector. 

\paragraph{Knowledge Fusion.}
The knowledge fusion problem involves how to combine the finetuned model sets $\{\theta_1, \hdots, \theta_n\}$ to form a new model $\theta_m$, without the need to retrain using the initial training data for each task, and ensuring that $\theta_m$ can simultaneously perform tasks $\{1, \hdots, n\}$. 
The multi-task model merging via task vectors is expressed as:
\begin{equation}
    \theta_m = \theta_\textrm{pre} + \sum\nolimits_{i=1}^{n}(\lambda_i \cdot m_i \odot {\tau}_i) / \sum\nolimits_{i=1}^{n}{\lambda}_i
\end{equation}
Here, \(\lambda_i\) is the coefficient for a specific pruned task vector, which can be optimized using evolutionary strategies to obtain an optimal set of \(\{\lambda_1, \dots, \lambda_n\}\) with the maximum validation accuracy for the final merged model.

\paragraph{Knowledge Compression.}
% Pruning fine-tuned models can be an effective strategy for compressing checkpoints. By utilizing sparsity masks, we can retain models' full performance while significantly reducing storage requirements.
Pruning fine-tuned models is an effective strategy for compressing checkpoints. By applying sparsity masks to weights and storing only the masked values, we can preserve full performance while greatly reducing storage.

% Instead of saving the entire collection of fine-tuned checkpoints
In term of storage for
\(\{\theta_t\}_{t=1}^T\), we only need to store the pre-trained model \(\theta_\textrm{pre}\), the task vectors \(\tau\), and the binary masks \(m\) for each task. For multi-task evaluation, fine-tuned models can be reconstructed by adding only the important subsets of task-specific vectors to the shared pretrained parameters\(\theta_\textrm{pre}\):

\begin{equation} 
\hat{\theta}_{\textrm{ft}_1}, \ldots, \hat{\theta}_{\textrm{ft}_n} = \theta_{\text{pre}} + [m_1 \odot \tau_1, \ldots, m_n \odot \tau_n]
\end{equation} 
% We provide the details for storage comparison in Appendix~\ref{app:storage}.

\section{Experiment}
\label{sec:exp}

  \begin{table*}[t]
    \caption{Average performance and H-score on LLaVA-1.5 (Vicuna-7B) with a sparsity ratio $r=10\%$. ``\#Params" refers to the number of parameters modified. The optimal and sub-optimal results are denoted by boldface and underlining. \looseness=-1}
    \vspace{-3mm}
  \label{tab:main_llava}
  \centering
  % \scriptsize
  \resizebox{1\linewidth}{!}{
  \begin{tabular}{r|l|cccccccc|c|>{\columncolor{tabhighlight}}c >{\columncolor{tabhighlight}}c}

      \bottomrule
      
      \toprule
      \multirow{2}{*}{Method} &  \multirow{2}{*}{\#Params} &  \multicolumn{8}{c}{\textbf{Pre-trained tasks}} & \multicolumn{1}{c}{\textbf{Target task}}\\
      & &
      \multicolumn{1}{c}{\textcolor{azureblue}{VQAv2}} &
      \multicolumn{1}{c}{\textcolor{azureblue}{GQA}} &
      \multicolumn{1}{c}{\textcolor{azureblue}{VizWiz}} &
      \multicolumn{1}{c}{\textcolor{azureblue}{SQA}} &
      \multicolumn{1}{c}{\textcolor{azureblue}{TextVQA}} &
      \multicolumn{1}{c}{\textcolor{azureblue}{POPE}} &
      % \multicolumn{1}{c}{MME} &
      \multicolumn{1}{c}{\textcolor{azureblue}{MM-Bench}} &
       \multicolumn{1}{c}{\textcolor{azureblue}{MM-Bench-CN}} &
        % \multicolumn{1}{c}{MM-Vet} &
      \multicolumn{1}{c}{\textcolor[rgb]{0.63,0.113,0.167}{Flickr30k}} &
      \multicolumn{1}{c}{Avg} &
      \multicolumn{1}{c}{Hscore} 
      \\  
      
      \midrule
      {Zero-shot}  & -& 78.52 &  61.97 &  50.0 & 70.17 & 58.28 & 85.97 & 64.78 & 58.51 & 18.62 & 42.33 & 29.05      \\
    \midrule
  % {DARE-Lora}  &343M  & & & & & & & & & &    &  &  &           \\
  % {Grafting-Lora}  & 343M & & & & & & & &  &  &  &   &  &               \\
  % \midrule
  {Fine-tune} & 2.7B & 68.61 & 49.01 &27.24 & 63.86 &40.03 & 79.73& 59.02&50.17 &  77.1 & 56.42 &63.40      \\
  % {LoRA} & 343M & 76.35 & 58.64&50.25 &65.13 & 55.18 & 84.91& 64.43&53.09 & 62.5 & 63.39 &62.99       \\
  {DARE\pub{ICML24}}  & 273M& {78.12} & {59.25}& {48.9}&64.92 & {57.17}& {84.86}&{64.77}& {57.47}& 25.6 & 60.12 & 36.64           \\
  {Grafting\pub{ICML23}}  & 273M& {74.48} &{58.28} &{43.16} &{66.82} &{52.56} &80.35& {64.52} & 55.49 &{58.2}  & 61.56 & 60.03             \\
  % \rowcolor{tabhighlight}
 {Model Tailor\pub{ICML24}}   & 273M&73.21 &52.49 &42.28 & {67.15} &43.89 & {82.88}& 63.40 & {56.15} & {75.4}  & \underline{61.87} &\underline{66.94}         \\
  \textbf{\ourapproach (ours)}   & 273M & 74.3 & 52.52 & 43.1 & 66.12 & 43.93 & 83.23 &  64.52 & 57.51 & 76.2 &  \textbf{62.38} & \textbf{67.54}       \\
  % \rowcolor{tabhighlight}
  % {Ours-Lora}  & 343M & & & & & & & & &  &  & &  &    \\
  \bottomrule

  \toprule

   \multirow{2}{*}{Method} & \multirow{2}{*}{\#Params} &  \multicolumn{8}{c}{\textbf{Pre-trained tasks}} & \multicolumn{1}{c}{\textbf{Target task}}\\
      % \cline{2-14}
      % \midrule
      & &
       \multicolumn{1}{c}{\textcolor{azureblue}{VQAv2}} &
      \multicolumn{1}{c}{\textcolor{azureblue}{GQA}} &
      \multicolumn{1}{c}{\textcolor{azureblue}{VizWiz}} &
      \multicolumn{1}{c}{\textcolor{azureblue}{SQA}} &
      \multicolumn{1}{c}{\textcolor{azureblue}{TextVQA}} &
      \multicolumn{1}{c}{\textcolor{azureblue}{POPE}} &
      % \multicolumn{1}{c}{MME} &
      \multicolumn{1}{c}{\textcolor{azureblue}{MM-Bench}} &
       \multicolumn{1}{c}{\textcolor{azureblue}{MM-Bench-CN}} &
        % \multicolumn{1}{c}{MM-Vet} &
      \multicolumn{1}{c}{\textcolor[rgb]{0.63,0.113,0.167}{OKVQA}} &
      \multicolumn{1}{c}{Avg} &
      \multicolumn{1}{c}{Hscore} 
      \\  
      
      \midrule
      {Zero-shot}   & -& 78.52 &  61.97 &  50.0 & 70.17 & 58.28 & 85.97 & 64.78 & 58.51 & 0.14 & 27.94 & 33.09       \\
      \midrule
  {Fine-tune} &2.7B & 69.1& 48.61&30.35 &41.03 &42.13 & 72.33 &32.79 &43.47 & 46.27& 47.34 &46.87     \\
  % {LoRA} &343M &71.21 & 52.87&37.09 &28.15 &54.36 & 67.81& 35.91 &29.47 &59.25 &48.45  &52.48      \\
  {DARE\pub{ICML24}}  & 273M& {78.04} &  {61.65} & {49.19} & {67.58}& {57.91} & {86.44}  & {65.03}& {58.16}& 0.83 & 58.31 & 1.64           \\
  {Grafting\pub{ICML23}}  & 273M&75.23 & 58.42& 43.27&67.26 &53.51 &85.29  & 62.16& 54.42& 30.8 & {58.93}& {41.25}              \\
  % \rowcolor{tabhighlight}
  {Model Tailor}\pub{ICML24}   & 273M&  {76.25} & {60.39}&  {46.49}& {69.51} & {54.88} & {85.44} & {63.32}& 54.21&  {38.1} & \underline{60.95}  &\underline{47.71}   \\
  \textbf{\ourapproach (ours)}   & 273M&  76.81 & 60.94 & 48.1 & 71.32 & 56.34 & 87.23 &  64.77 & 57.5 & 38.4 & \textbf{62.38} & \textbf{48.38}  \\
  \bottomrule

  \toprule

  \end{tabular}
  }
  % \vspace{-5mm}
  \end{table*}

\subsection{Evaluation Settings}
We expect that \ourapproach will provide significant benefits for developers in three main areas:
First, in experiments with multimodal large language models (MLLMs) using the LLaVA framework, our approach preserved performance even at a sparsity level of 10\%. This highlights its effectiveness in mitigating catastrophic forgetting. Second, in knowledge fusion, it consistently outperforms existing merging techniques across various modalities, domains, and model sizes. Lastly, for knowledge compression, it achieves superior accuracy and storage efficiency compared to baselines on ViT-based vision tasks. 
% First, it effectively mitigates catastrophic forgetting in knowledge transfer scenarios. In experiments with multimodal large language models (MLLMs) using the LLaVA framework, our approach preserved performance even at a sparsity level of 10\%. This highlights its effectiveness in mitigating catastrophic forgetting.
% Second, in knowledge fusion, \ourapproach has been evaluated across various scenarios, including different modalities, domains, model sizes, fine-tuning methods, and large language models. It consistently outperforms existing model merging techniques.
% Lastly, for knowledge compression, we compared our method against baselines by evaluating both accuracy and storage cost across different task combinations on vision benchmarks using ViT models, where our approach demonstrated superior performance. 
More information on implementation details can be found in Appendix~\ref{app:c}.

\subsection{Baseline Methods}
Our baselines are categorized into three primary areas: knowledge transfer for mitigating catastrophic forgetting, knowledge fusion, and compression. For knowledge transfer, we compare our approach against Standard \textbf{Fine-tuning}, \textbf{Model Grafting}~\cite{panigrahi2023task}, Drop \& Rescale (\textbf{DARE})~\cite{yu2023language}, and \textbf{Model Tailor}~\cite{zhumodel}. In the domain of knowledge fusion, we assess various methods such as \textbf{Simple Averaging}~\cite{wortsman2022model}, \textbf{Fisher Merging }\cite{matena2022merging}, \textbf{RegMean}~\cite{jin2023regmean}, \textbf{Task Arithmetic }\cite{ilharco2022editing}, \textbf{Ties-Merging}~\cite{ties}, \textbf{PCB Merging}~\cite{pcb} and \textbf{Consensus Merging}~\cite{talls}. Notably, Task Arithmetic, Ties-Merging, Consensus Merging, and our proposed \ourapproach are all based on task vectors, making them training-free and lightweight. For knowledge compression, we evaluate our method against several model merging techniques and their combinations with \textbf{Talls Mask}~\cite{talls}. Detailed information on these baselines can be found in Appendix~\ref{app:d}.

\subsection{Results on Knowledge Transfer}
Following ~\cite{liu2023improved}, we conduct knowledge transfer experiments using LLaVA-1.5 (Vicuna-7B). 
Both the projector and LLM parameters of the model are fine-tuned. 
The pre-trained datasets include VQAv2~\cite{goyal2017making}, GQA~\cite{hudson2019gqa}, Vizwiz~\cite{gurari2018vizwiz}, SQA~\cite{lu2022learn}, TextVQA~\cite{singh2019towards}, POPE~\cite{li2023evaluating}, MM-Bench~\cite{liu2023mmbench}, and MM-Bench-CN~\cite{zhang2023internlm}.
We then fine-tune LLaVA on Flickr30k~\cite{young2014image} and OKVQA~\cite{marino2019ok} tasks, which are not included in the model's pre-training datasets. 
The performance of the fine-tuned model is evaluated on these and other datasets.

For evaluation, we use both the arithmetic and harmonic means~\cite{zhumodel} of performance across pre-trained and target tasks, referred to as average performance and H-score. As shown in Table \ref{tab:main_llava}, our \ourapproach method effectively mitigates catastrophic forgetting in MLLMs, outperforming current fine-tuning and forgetting mitigation techniques at a sparsity level of 10\%. While further fine-tuning to improve performance on new tasks often deteriorates the model’s effectiveness on pre-trained tasks, \ourapproach successfully balances targeted optimization with the preservation of pre-trained performance. It achieves superior average metrics, improving by 1.5\% and 1.4\%, respectively, demonstrating its capability to enhance task-specific performance while maintaining robustness.

\begin{table*}[ht]
\captionsetup{type=table}
\caption{Comparison of different model merging methods across various fine-tuning configurations and modalities, with average performance reported for different tasks. The optimal and sub-optimal results are denoted by boldface and underlining. \looseness=-1}
\label{tab:merging} 
\centering
\resizebox{1.0\linewidth}{!}{  
\begin{tabular}{r|ll|l|l|ll|ll}
\thickhline
% \rowcolor{mygray}
Settings ($\rightarrow$)  & \multicolumn{2}{c|}{\small7 \textbf{NLP} Tasks} &\small 11 \textbf{PEFT} Tasks & \small3 \textbf{LLM} Tasks & \multicolumn{2}{c}{\small8 \textbf{Vision} Tasks} & \multicolumn{2}{|c}{\small5 \textbf{Emotion} Domains}\\ 
\cline{2-9}
% \cmidrule(lr){7-8}
% \hline
% \rowcolor{mygray}
Method ($\downarrow$) & T5-Base & T5-Large & ~~~~~~~(IA)$^3$ & ~~LLaMa2 & ViT-B/32 & ViT-L/14 & T5-Base & RoBERTa-Base \\

\hline
% \rowcolor{gray!20} 
Fine-tuned & 83.1 & 88.9 & 71.4 & 40.4 & 90.5 & 94.2 & 51.38 & 49.38 \\
Multitask  & 83.6 & 88.1 & 73.1 & -    & 88.9 & 93.5 & 47.75 & 49.06 \\
\hline
Averaging\pub{ICML22} & 65.3 & 54.7 & 57.9 & 30.3 & 65.8 & 79.6 & 23.2 & 38.3 \\ %\cite{wortsman2022model}
% Task Arithmetic\pub{ICLR23} \cite{ilharco2022editing}  & 53.5 & 73.6 & 59.2 & 30.4  & 60.4 & 83.3 \\
% Ties-Merging\pub{NeurIPS23} \cite{ties}  & 69.5 & 71.7  & 64.9 & 34.2 & 72.4 & 86.0  \\
% \textbf{\ourapproach (ours)}  & \textbf{73.8 \textcolor{color2}{(+4.3)}} & \textbf{77.1 \textcolor{color2}{(+3.5)}} & \textbf{66.1 \textcolor{color2}{(+1.2)}} & \textbf{35.1 \textcolor{color2}{(+0.9)}} & \textbf{75.9 \textcolor{color2}{(+3.5)}} & \textbf{86.9 \textcolor{color2}{(+0.9)}}  \\ 
% \hline
Fisher Merging\pub{NeurIPS22}   & 68.3 & 68.7  & 62.2 & -  & 68.3 & 82.2 & 26.1 & 38.1 \\ %\cite{matena2022merging}
RegMean\pub{ICLR23}        & 72.7 & 79.8  & 58.0   & -  & 71.8 & 83.7 & 34.2 & 38.4 \\%\cite{jin2023regmean} 
Task Arithmetic\pub{ICLR23}   & 73.0 & 80.2  & 63.9 & 30.4 & 70.1 & 84.5 & 33.6 & 38.3 \\%\cite{ilharco2022editing}
Ties-Merging\pub{NeurIPS23}           & \underline{73.6} & 80.3  & \underline{66.8} & 34.2 & 73.6 & 86.0 & \underline{34.5} & 39.7 \\ 
%\cite{ties}
Consensus TA\pub{ICML24}  & 73.1 & 80.2  & 65.8 & 33.5 & \underline{73.5} & 85.8 & 33.9 & 39.2\\
%\cite{talls}
Consensus TIES\pub{ICML24} & 73.4 & \underline{80.5}  & 66.6 & \underline{34.4} & 73.3 & \underline{86.2} & 34.4 & \underline{39.8} \\
%\cite{talls}
\textbf{\ourapproach (ours)}    & \textbf{75.7 \textcolor{color2}{(+2.1)}} & \textbf{82.1 \textcolor{color2}{(+1.6)}}  & \textbf{68.2 \textcolor{color2}{(+1.4)}} & \textbf{35.3 \textcolor{color2}{(+0.9)}}   & \textbf{76.5 \textcolor{color2}{(+3.0)}} & \textbf{87.6 \textcolor{color2}{(+1.4)}} & \textbf{35.7 \textcolor{color2}{(+1.3)}} & \textbf{40.9 \textcolor{color2}{(+0.9)}} \\ 
% \textbf{\ourapproach+ ES (ours)}  & \textbf{76.7 \textcolor{color2}{(+3.1)}} & \textbf{83.2 \textcolor{color2}{(+2.9)}}  & \textbf{68.8 \textcolor{color2}{(+2.0)}} & \textbf{35.3 \textcolor{color2}{(+1.1)}}  & \textbf{77.0 \textcolor{color2}{(+3.4)}} & \textbf{88.1 \textcolor{color2}{(+2.1)}}  \\ 
\thickhline
\end{tabular}
}
\end{table*}

\subsection{Results on Knowledge Fusion}
    \begin{figure*}[t]
    \centering
    \includegraphics[width=0.9\textwidth]{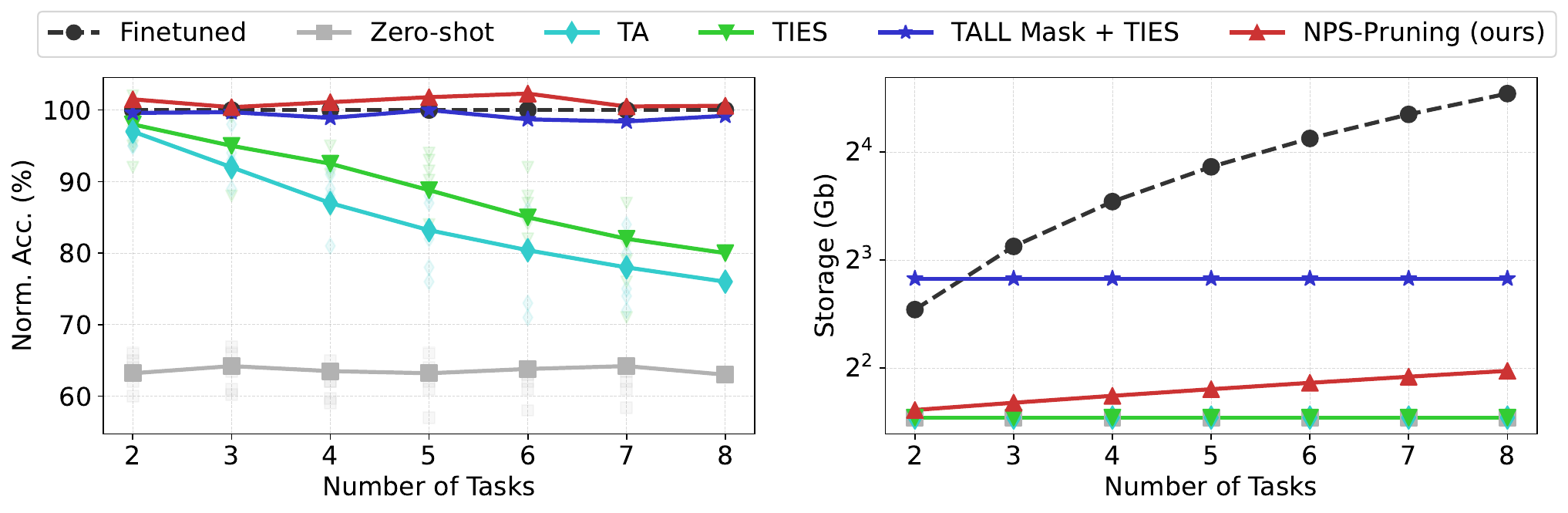}
    \caption{Averaged normalized accuracy and storage cost versus the number of tasks on computer vision benchmarks. Our proposed \ourapproach method consistently preserves initial performance across various task combinations while significantly compressing the fine-tuned checkpoints.}
    \label{fig:compression}
    \end{figure*}
To empirically validate the effectiveness of \ourapproach, we conducted extensive experiments to compare it with existing model merging techniques. Our results highlight the advantages of our approach across both cross-task and cross-domain perspectives. Detailed information about the datasets used is provided in Appendix~\ref{app:e}.

\paragraph{Merging NLP Models.}
In the NLP domain, we follow the experimental setup outlined in~\cite{ties}. We use the T5-base and T5-large models~\cite{raffel2020exploring}, fine-tuning each on seven diverse tasks, including question answering, paraphrase identification, sentence completion, and coreference resolution. Table \ref{tab:merging} demonstrates that merging fully fine-tuned T5-base and T5-large models using \ourapproach results in an average performance improvement of 2.1\% for T5-base and 1.6\% for T5-large across the seven tasks.

\paragraph{Merging PEFT Model Adapters.}
Based on~\cite{ties}, we explore parameter merging for efficient fine-tuning using the (IA)$^3$ method~\cite{liu2022few}, a type of Parameter-Efficient Fine-Tuning (PEFT) that extends base model activations with learned vectors. We use the T0-3B model~\cite{sanh2022multitask} and fine-tune (IA)$^3$ on training sets from eleven diverse datasets, including tasks such as sentence completion and natural language inference.
% We utilize prompt templates from the Public Prompt Pool (P3~\cite{bach2022promptsource}) to convert dataset examples into a text-to-text format, with each label as a different string. 
For the (IA)$^3$ experiments, we report median scores across all templates for each dataset. As shown in Table~\ref{tab:merging}, \ourapproach improves average performance by 1.4\% across 11 tasks compared to the top baseline.

\paragraph{Merging LLMs.}
In our experiment, we combined three specialized large language models built on the Llama-2-7b architecture~\cite{touvron2023llama}, each focusing on a different area: Chinese language proficiency\footnote{\url{https://huggingface.co/LinkSoul/Chinese-Llama-2-7b}}, mathematical reasoning~\cite{yu2024metamath}\footnote{\url{https://huggingface.co/meta-math/MetaMath-7B-V1.0}}, and code generation~\cite{rozière2024code}\footnote{\url{https://huggingface.co/qualis2006/llama-2-7b-int4-python-code-18k}}. We assessed the performance of each model using specific benchmarks: CMMLU~\cite{li2024cmmlu} for Chinese, GSM8K~\cite{cobbe2021training} for mathematics, and HumanEval~\cite{chen2021evaluating} for code generation. As indicated in Table~\ref{tab:merging}, our method \ourapproach resulted in an average performance improvement of 0.9\%.

\paragraph{Merging Vision Models.}
For image classification tasks, we adhered to the experimental setup outlined by~\cite{ilharco2022patching, ilharco2022editing}. We employed two versions of the CLIP model~\cite{radford2021learning}, specifically using ViT-B/32 and ViT-L/14 as visual encoders. The visual encoders were fine-tuned on eight tasks sourced from~\cite{radford2021learning}, while the text encoder remained unchanged. This approach covered a range of classification domains, such as remote sensing, traffic classification, and satellite imagery recognition. Our method achieved a 3.0\% improvement over the top baseline on ViT-B/32 and a 1.4\% improvement on ViT-L/14.

\paragraph{Merging Emotion Domains.}
We carried out further experiments to evaluate the effectiveness of various methods in merging five domain-specific emotion classification models. In line with the methodology of RegMean~\cite{jin2023regmean}, we used the Roberta-base and T5-base models, along with five preprocessed datasets from~\cite{oberlander2018analysis}. Our analysis presents the average accuracy on in-domain datasets achieved by different model merging techniques. Additionally, we conducted experiments with multiple random seeds and reported the average results across five seeds. As detailed in Table \ref{tab:merging}, our approach surpasses the best baseline by 1.3\% on Roberta-base and 0.9\% on T5-base.

\subsection{Results on Knowledge Compression}
We conducted experiments using eight different ViT-B/32 models, each fine-tuned on distinct vision tasks, and tested the performance and compression efficiency across various numbers of tasks. For each task quantity, five random combinations were selected, and the average results were reported.
As shown in Figure 8, both TALL-Mask and \ourapproach maintain around 99\% normalized accuracy across all cases, with virtually no performance degradation as the number of tasks increases.

In terms of storage, our method significantly reduces costs compared to storing individual fine-tuned models, with the savings becoming more pronounced as the number of tasks increases.
The TALL Mask + TIES method consistently consumes a high amount of storage, even when the number of tasks is small. In contrast, our approach requires storage that increases gradually with the number of tasks. While methods like Task Arithmetic have lower storage demands, they suffer from a noticeable drop in accuracy.
Overall, our method achieves an optimal balance on the Pareto front, effectively retaining performance while minimizing total storage costs.
More results about knowledge compression are provided in supplemental materials Appendix~\ref{app:a}.

\section{Analysis}
\label{sec:analysis}
% \begin{figure*}[b]
% \centering
% \includegraphics[width=\textwidth]{figures/compression.pdf}
% \caption{The performance and storage of different methods.}
% \label{fig:compression}
    % \end{figure*}
    \begin{figure}[t]
    \centering
    \includegraphics[width=0.43\textwidth]{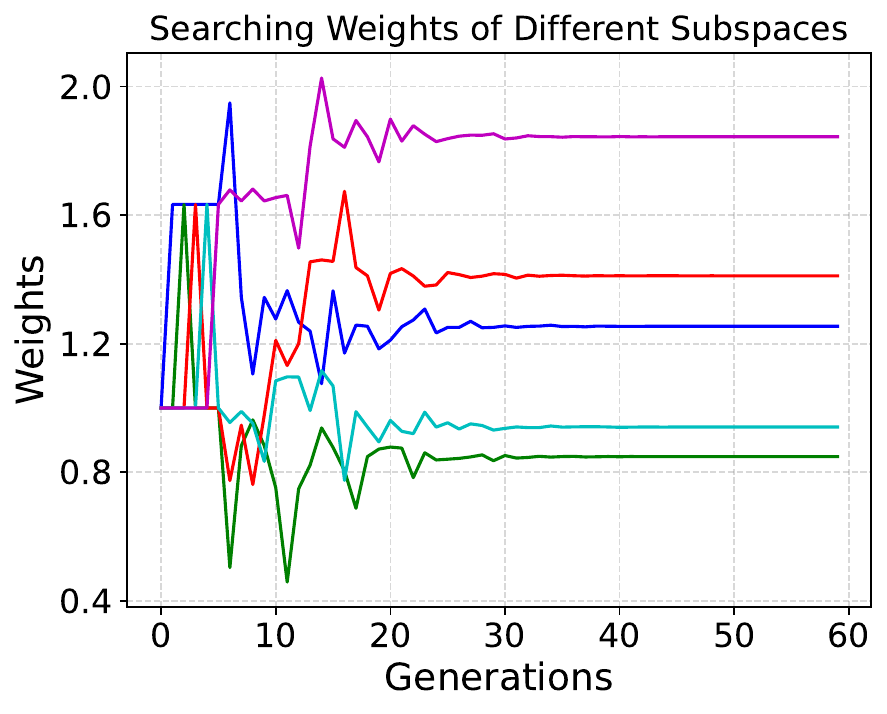}
    \caption{Searching for the weights of neural parameters across different task vector subspaces.}
    \label{fig:process1}
    \vspace{-4mm}
    \end{figure}  
  
\begin{figure}[t] 
    \centering
\includegraphics[width=0.438\textwidth]{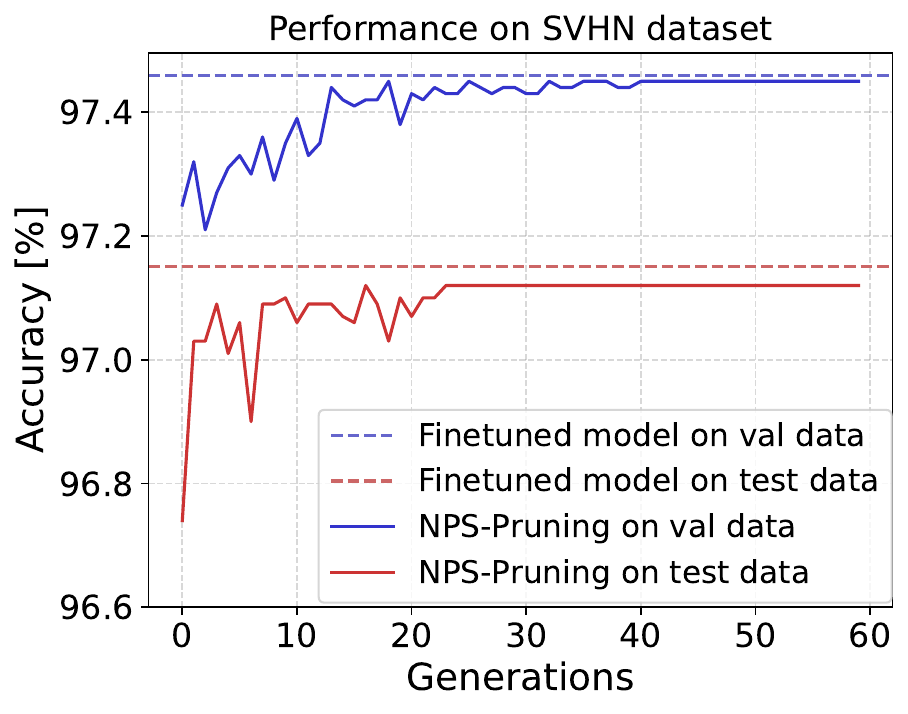}
    \caption{Performance convergence of the pruned fine-tuned model as the number of generations increases.}
    \label{fig:process2}
    \vspace{-3mm}
    \end{figure}

\paragraph{Search Visualization.}
To better understand the workflow of our method, we visualized the pruning process for a ViT-B/32 model fine-tuned on the SVHN dataset, setting the sparsity ratio to 0.1.  
We divided the task vector into five subspaces based on their magnitude values and then continuously updated the weights of these subspaces to explore higher validation scores. 
It can be observed that the weight values stabilize as the number of generations increases, as shown in Figure~\ref{fig:process1}, and the pruned model's accuracy also gradually converges to a stable value, as shown in Figure~\ref{fig:process2}.

\paragraph{Time complexity.}
The total time required for the overall \ourapproach strategy is 
\begin{equation}
    T_{\text{total}} = \text{Generations} \times (T_{\text{pruning}} + T_{\text{validate}})
\end{equation} 
where generations represents the number of generations needed for searching, which is a pre-set value and varies with different experiment settings. 
The pruning time mainly depends on the number of model parameters and the size of the model population, while the validation time primarily depends on the volume of inference data and the inference speed.  We have organized ablation study and reported the number of generations and time required in our experiments, as shown in Appendix~\ref{app:b}.
    
\paragraph{Advantages.}
Our method offers several notable benefits, making it an efficient, flexible, and practical solution for various use cases.
\begin{itemize}
    \item \textbf{Gradient-Free Operation}: \ourapproach method operates without gradient calculations, making it lightweight and minimizing memory usage. 
    % This is particularly advantageous in environments with limited computational resources.
    \item \textbf{Practicality and Ease of Implementation}: The method is straightforward to implement and integrates easily into various applications. 
    % Its simplicity enhances its usability across different tasks and scenarios.
    \item \textbf{Broader Applicability and Stable Performance}: Unlike theoretical pruning methods, our approach is more versatile and provides consistent results across various applications. 
    % Its stability makes it a reliable choice for practical use.
\end{itemize}
\section{Conclusions}
\label{sec:conclusions}
This study highlights the significance of pruning fine-tuned models when pretrained model is available. We introduce Neural Parameter Search (NPS) as an efficient technique for this task. Our approach facilitates multi-task model fusion, compression, and robust knowledge transfer by searching neural parameters within task vector subspaces. Experimental results demonstrate that \ourapproach significantly enhances performance across various knowledge transfer scenarios.

\section*{Acknowledgements}
This work was supported by National Science Foundation of China (62476070), Shenzhen Science and Technology Program \seqsplit{(JCYJ20241202123503005,~ GXWD20231128103232001,~ ZDSYS20230626091203008,~ KQTD2024072910215406)}  and Department of Science and Technology of Guangdong (2024A1515011540).

% \newpage
\section*{Limitations}
While our method offers advantages such as gradient-free operation, ease of implementation, and broad applicability, it also has certain limitations:  
\begin{itemize}
\item \textbf{Dependence on Pretrained Models}: Our approach relies on pretrained models as a reference. If the fine-tuned model deviates significantly from the original, it may hinder effective knowledge transfer, fusion, and compression.  
\item \textbf{Validation Data Requirements}: The method requires additional validation data to guide the search process. The quality and quantity of this data directly impact pruning effectiveness and overall performance.  
\item \textbf{Computational Overhead of the Search Process}: Although the method is gradient-free, the search process introduces a time cost, which varies depending on task complexity. This trade-off should be considered when deploying the method in resource-constrained environments.  
\end{itemize}
% We have focused exclusively on four commonly used modalities, leaving out a thorough analysis of the full range of potential modalities.
% Additionally, finding multiple existing MLLMs with the same architecture across modalities is currently challenging, and due to limited computational resources, experiments on larger-scale MLLMs are constrained.
% Finally, although our \ourapproach approach does not increase inference parameters, the storage cost is twice that of the base model.

\section*{Ethical Considerations}
Our research is based on publicly available and safe datasets and models. However, the applicability of \ourapproach may be limited to similar datasets or domains. Its performance on other datasets remains uncertain, and applying it to privacy-sensitive or high-risk scenarios may pose risks. We recommend caution and thorough validation to ensure accuracy and reliability in such cases.
\newpage
\bibliography{custom}

\newpage
\appendix
\appendix
% \textbf{{\Large Appendix for \ourapproach{}}}
% \title{\Large Appendix for \ourapproach{}}
\title{Appendix for Neural Parameter Search For Knowledge Transfer, Fusion and Compression}
% \title{Appendix for NPS-Pruning}
\maketitle
~~~~~~~~~~~~~~~~~~~~~~~~~~~~~~~ \textbf{Appendix} \\
This paper enhances the pruning efficiency of fine-tuned models through \textbf{N}eural \textbf{P}arameter \textbf{S}earch and applies this approach to various scenarios, including knowledge transfer, fusion, and compression, with the assistance of pre-trained models. The appendix is organized based on the following contributions:
\begin{itemize}
    \item Appendix~\ref{app:a} (Additional Results) provides additional experimental results on knowledge compression as well as task-level results from the knowledge fusion experiments.
    \item Appendix~\ref{app:b} (Additional Analysis) includes ablation studies, hyperparameter analysis, and time cost evaluation for the search process.
    \item Appendix~\ref{app:c} (Implementation Details) outlines the computational resources and runtimes, along with the training details and evaluation metrics.
    \item Appendix~\ref{app:d} (Baselines) provides a detailed baseline description.
    \item Appendix~\ref{app:e} (Datasets) provides a detailed dataset description.
\end{itemize}
\begin{table*}[th]
\captionsetup{type=table}
\caption{Comparison of different knowledge compression methods across various modalities, with average performance reported for different tasks. The optimal results are denoted by boldface. Please refer to Section 4.5 for more
details. \looseness=-1}
\label{tab:app_compression} 
\centering
\resizebox{1.0\linewidth}{!}{  
\begin{tabular}{r|cc|cc|cc|cc|cc}
\thickhline
% \rowcolor{mygray}
Settings ($\rightarrow$)  & \multicolumn{4}{c|}{\small7 \textbf{NLP} Tasks} & \multicolumn{2}{c|}{\small3 \textbf{LLM} Tasks} & \multicolumn{4}{c}{\small8 \textbf{Vision} Tasks} \\ 
\cline{2-11}
% \cmidrule(lr){7-8}
% \hline
% \rowcolor{mygray}
 & \multicolumn{2}{c|}{T5-Base} & \multicolumn{2}{c|}{T5-Large} & \multicolumn{2}{c|}{LLaMa2} & \multicolumn{2}{c|}{ViT-B/32} & \multicolumn{2}{c}{ViT-L/14} \\
Method ($\downarrow$) & Acc.(\%)$\uparrow$ & Bits(Gb)$\downarrow$ & Acc.(\%)$\uparrow$ & Bits(Gb)$\downarrow$ & Acc.(\%)$\uparrow$ & Bits(Gb)$\downarrow$ & Acc.(\%)$\uparrow$ & Bits(Gb)$\downarrow$ & Acc.(\%)$\uparrow$ & Bits(Gb)$\downarrow$ \\
\hline
\rowcolor{gray!20} 
Fine-tuned & 83.1& 47.8 & 88.9 & 169.1 & 40.4 & 629.6 & 90.5 & 23.3 & 94.2 & 79.1 \\
\rowcolor{gray!20} 
Zero-shot  & 53.5$_{(64.4)}$ & 7.1 & 53.1$_{(59.7)}$ & 25.1 & 15.3$_{(37.9)}$ & 215.6 & 62.3$_{(68.8)}$ & 3.6 & 74.5$_{(79.1)}$ & 11.0 \\
\hline
% Averaging\pub{ICML22} & 65.3 & 54.7 & 57.9 & 30.3 & 65.8 & 79.6 & 23.2 & 38.3 \\ 
% \hline
Task Arithmetic\pub{ICLR23}   & 73.0$_{(87.8)}$ & 7.1  & 80.2$_{(90.2)}$ & 25.1 & 30.4$_{(75.2)}$ & 215.6 & 70.1$_{(77.5)}$ & 3.6 & 84.5$_{(89.7)}$ & 11.0 \\%\cite{ilharco2022editing}
TIES\pub{NeurIPS23}           & 73.6$_{(88.6)}$ & 7.1  & 80.3$_{(90.3)}$ & 25.1 & 34.2$_{(84.7)}$ & 215.6 & 73.6$_{(81.3)}$ & 3.6 & 86.0$_{(91.3)}$ & 11.0 \\ 
Talls+TIES\pub{ICML24} & 82.6$_{(99.4)}$ & 15.2 & 88.3$_{(99.3)}$ & 54.3 & 39.5$_{(97.8)}$ & 442.3 & 90.2$_{(99.7)}$ & 7.1 & 93.6$_{(99.4)}$ & 23.1 \\
\textbf{\ourapproach (ours)} & \textbf{82.9}$_{(\textbf{99.8})}$ & 11.1 & \textbf{88.8}$_{(\textbf{99.9})}$ & 39.2 & \textbf{40.5}$_{(\textbf{100.2})}$ & 276.3 & \textbf{90.9}$_{(\textbf{100.4})}$ & 5.9 & \textbf{94.3}$_{(\textbf{100.1})}$ & 18.0 \\

\thickhline
\end{tabular}
}
\end{table*}

\section{Additional Results}
\label{app:a}
\subsection{Additional Results on Compression}
In our NLP experiments, particularly in the knowledge compression scenarios involving large language models, we present additional results, as shown in Appendix Tables~\ref{tab:app_compression}. These results demonstrate that our method maintains the performance of the previous best compression approach, TALLS Mask+TIES, while significantly reducing storage consumption.
\subsection{Comprehensive Task-Level Results}
We present task-level results for all knowledge fusion experiments in Section 4.4. Detailed task-level outcomes for T5-Base, T5-Large \cite{raffel2020exploring}, IA3 \cite{liu2022few}, ViT-B/32, and ViT-L/14 \cite{dosovitskiy2021an} are provided in Appendix Tables~\ref{tab:app_t5_base},~\ref{tab:app_t5_large},~\ref{tab:app_ia3},~\ref{tab:app_vit_base}, and~\ref{tab:app_vit_large}, respectively. We also provide radar charts to compare the results of merging vision tasks, as illustrated in Appendix Figure~\ref{fig:vision_tasks}. While previous baseline methods exhibit inconsistent performance and struggle with certain tasks, our method proves to be more robust, delivering near-optimal results across all tasks.
\begin{figure*}[thbp]
    \centering
    \includegraphics[width=0.9\linewidth]{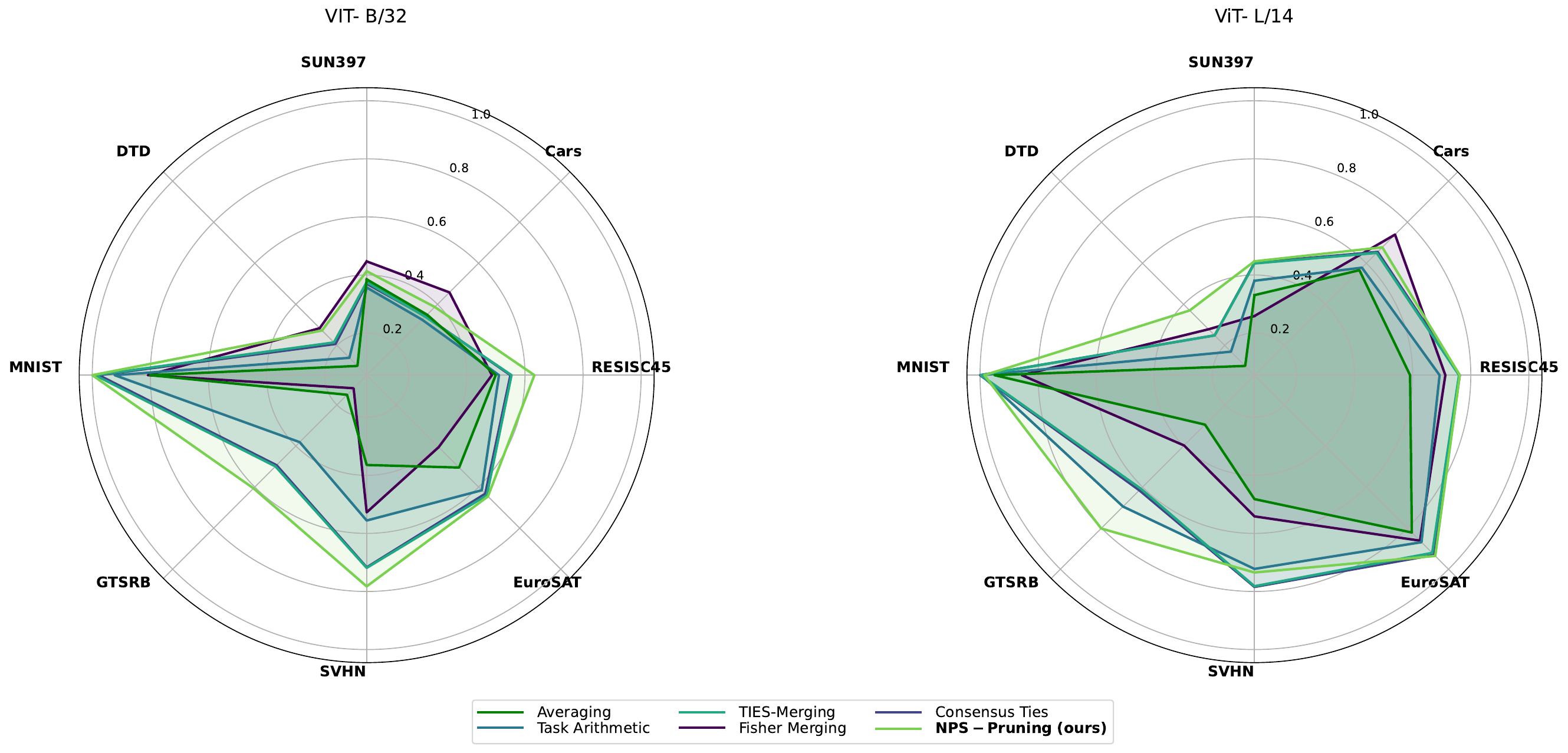}
    \captionsetup{type=figure}
    \caption{Test set performance when merging ViT-B/32 and ViT-L/14 models on eight image classification tasks.}
    \label{fig:vision_tasks}
\end{figure*}

\begin{table*}[htb!]
\caption{Test set performance when merging T5-base models on seven NLP tasks. Please refer to Section 4.4 for more details.}
\label{tab:app_t5_base}
% \vspace{10pt}
\centering
\belowrulesep=0pt
\aboverulesep=0pt
\resizebox{0.9\linewidth}{!}{  
\begin{tabular}{r|c|ccccccc}
\toprule
\rowcolor{mygray}\textbf{Task($\rightarrow$)} &   & \multicolumn{7}{c}{\textbf{Test Set Performance}} \\ 
\cline{3-9}
\rowcolor{mygray}\textbf{Method($\downarrow$)} & \multirow{-2}{*}{\textbf{Average}}  & paws  &  qasc  &  quartz  &  story\_cloze  &  wiki\_qa  &  winogrande  &  wsc \\
\midrule
\textbf{Zeroshot} & 53.5   &  49.9  &  35.8  &  53.3  &  48.1  &  76.2  &  50  &  61.1 \\
\textbf{Fine-tuned} & 83.1  &  94.6  &  98.4  &  81.1  &  84.9  &  95.8  &  64.5  &  62.5 \\
\textbf{Multitask} &  83.6 &  94  &  97.9  &  82.5  &  86.7  &  95  &  64.1  &  65.3 \\
\midrule
\textbf{Averaging}\pub{ICML22} & 65.3  &  67.4  &  83.4  &  60.8  &  50.3  &  93.2  &  51.7  &  50.0 \\
\textbf{Fisher Merging}\pub{NeurIPS22}  &  68.3 & 66.7 & 85.6 & 63.5 & 57.1 & 90.1 & 54.2 & 60.8\\
\textbf{RegMean}\pub{ICLR23} & 72.7 & 77.2 & \textbf{93.8} & 63.6 & 64.6 & 90.4 & 58.4 & 60.7 \\
\textbf{Task Arithmetic}\pub{ICLR23} & 73.0 & 69.6 & 91.5 & \textbf{67.3} & 76.1 & 91.3 & 58.3 & 56.9 \\
\textbf{Ties-Merging}\pub{NeurIPS23} & 73.6 & 82.2 & 84.8 & 66.1 & 73.5 & 87.0 & 60.2 & 61.1 \\
\textbf{Consensus Ties}\pub{NeurIPS23} & 73.4 & \textbf{82.3} & 84.5 & 65.7 & 73.4 & 86.8 & \textbf{60.3} & 60.5 \\
\textbf{\ourapproach (ours)} & \textbf{75.6} & 79.1 & 93.3 & 65.9 & \textbf{76.2} & \textbf{89.9} & 59.9 & \textbf{63.9} \\

\bottomrule
\end{tabular}
}
\vspace{10pt}
\end{table*}

% \begin{table*}[tbh!]
% \centering

% \resizebox{\linewidth}{!}{  
% \begin{tabular}{lccccccccc}
% \toprule

% \textbf{Method}  & \textbf{Validation} & \textbf{Average}  &  \textbf{paws}  &  \textbf{qasc}  &  \textbf{quartz}  &  \textbf{story\_cloze}  &  \textbf{wiki\_qa}  &  \textbf{winogrande}  &  \textbf{wsc} \\
% \midrule
% \textbf{Zeroshot}  & 51.7	& 55.4	& 14.3	& 54.1	& 54.3	& 70.9	& 49.2	& 63.9 \\
% \textbf{Fine-tuned} & 88.9	& 94.5	& 98.3	& 88.5	& 91.4	& 96.2	& 74.5	& 79.2 \\
% \textbf{Multitask} & 88.1 & 94.2 & 98.5 & 89.3 & 92 & 95.4 & 73.5 & 73.6 \\
% \midrule
% \textbf{Averaging} & \redxmark  &  54.7	& 57.2	& 26.4	& 71.4	& 54.8	& 86.6	& 50.2	& 36.1 \\
% \textbf{Task Arithmetic} & \redxmark & 73.6	& 69.7	& 83.6	& 58.3	& 77.4	& 94.4	& 59.3	& 72.2 \\
% \textbf{\methodshort{}} & \redxmark & 74.4 & 80.5 & 96.2 & 81.8 & 78.6 & 62 & 61.9 & 59.7 \\
% \midrule
% \textbf{Fisher Merging} & \greencmark  & 68.7	& 68.4	& 83	& 65.5	& 62.4	& 94.1	& 58.2	& 49.2 \\
% \textbf{RegMean} & \greencmark  & 79.8	& 83.9	& 97.2	& 73.2	& 82.6	& 94.1	& 63.2	 & 64.4 \\
% \textbf{Task Arithmetic} & \greencmark & 80.2	& 77.6	& 96.6	& 75.1	& 85.6	& 93.8	& 61.8	& 70.8 \\
% \textbf{Ties-Merging} & \greencmark &  80.3	& 78.2	& 97.5	& 72.8	& 83.7	& 94.5	& 64.5	& 70.8 \\
% \textbf{\methodshort{}} & \greencmark & & & & & & & & \\

% \bottomrule
% \end{tabular}
% }
% \caption{\label{tab:app_main_t5large} Test set performance when merging T5-large models on seven tasks. Please refer to Section \ref{sec:main_results} for experimental details.}
% \vspace{10pt}
% \end{table*}

\begin{table*}[htb!]
\centering
\belowrulesep=0pt
\aboverulesep=0pt
\caption{Test set performance when merging T5-large models on seven NLP tasks. Please refer to Section 4.4 for more details.}
\label{tab:app_t5_large} 
\resizebox{0.8\linewidth}{!}{  
\begin{tabular}{r|c|ccccccc}
\toprule
\rowcolor{mygray}\textbf{Task($\rightarrow$)}  &   & \multicolumn{7}{c}{\textbf{Test Set Performance}} \\
\cline{3-9}
\rowcolor{mygray}\textbf{Method($\downarrow$)} & \multirow{-2}{*}{\textbf{Average}} & paws  &  qasc  &  quartz  &  story\_cloze  &  wiki\_qa  &  winogrande  &  wsc \\
\midrule
\textbf{Zeroshot}  & 53.1	& 58.2	& 54.2	& 54.1	& 54.3	& 70.9	& 49.2	& 63.9 \\
\textbf{Fine-tuned} & 88.9	& 94.5	& 98.3	& 88.5	& 91.4	& 96.2	& 74.5	& 79.2 \\
\textbf{Multitask} & 88.1 & 94.2 & 98.5 & 89.3 & 92 & 95.4 & 73.5 & 73.6 \\
\midrule
\textbf{Averaging}\pub{ICML22}  &  54.7	& 57.2	& 26.4	& 71.4	& 54.8	& 86.6	& 50.2	& 36.1 \\
\textbf{Fisher Merging}\pub{NeurIPS22}  & 68.7	& 68.4	& 83	& 65.5	& 62.4	& 94.1	& 58.2	& 49.2 \\
\textbf{RegMean}\pub{ICLR23}  & 79.8	& \textbf{83.9}	& 97.2	& 73.2	& 82.6	& 94.1	& 63.2	 & 64.4 \\
\textbf{Task Arithmetic}\pub{ICLR23} & 80.2	& 77.6	& 96.6	& \textbf{75.1}	& 85.6	& 93.8	& 61.8	& 70.8 \\
\textbf{Ties-Merging}\pub{NeurIPS23} &  80.3	& 78.2	& 97.5	& 72.8	& 83.7	& \textbf{94.5}	& 64.5	& 70.8 \\
\textbf{Consensus Ties}\pub{NeurIPS23} &  80.5	& 78.4	& 97.7	& 72.6	& 83.7	& \textbf{94.8}	& 64.6	& 71.2 \\
\textbf{\ourapproach (ours)} & \textbf{82.1} & 82.1 & \textbf{98.4} & 72.3 & \textbf{85.7} & 94.1 & \textbf{67.2} & \textbf{75.0} \\

\bottomrule
\end{tabular}

}
\vspace{10pt}
\end{table*}

\begin{table*}[htb!]
\centering
% \belowrulesep=0pt
% \aboverulesep=0pt
\caption{Test set performance when merging (IA)$^3$ models on eleven tasks. Please refer to Section 4.4 for experimental details.}
\label{tab:app_ia3}
\resizebox{0.95\linewidth}{!}{  
\begin{tabular}{r|c|cccccc ccc cc c}

\toprule
\rowcolor{mygray}\textbf{Task($\rightarrow$)}  &   & \multicolumn{5}{c}{\textbf{Natural Language Inference}} & \multicolumn{3}{|c}{\textbf{Sentence Completion}} & \multicolumn{2}{|c}{\textbf{Co-reference}} & \multicolumn{1}{|c}{\textbf{WSD}}\\ 
\cline{4-13}

\rowcolor{mygray}\textbf{Method($\downarrow$)} & \multirow{-2}{*}{\textbf{Average}}  & RTE  &  CB  &  ANLI1  &  ANLI2  &  ANLI3  &  \multicolumn{1}{|c}{COPA}  &  Hella. & Story. & \multicolumn{1}{|c}{WSC} & Wino. & \multicolumn{1}{|c}{WiC}\\
\midrule

\textbf{Zeroshot}  & 53.1	& 58.2 & 54.2	& 35.5	& 34.4	& 34.4	& 75.0	& 39.2	& 86.5  & 63.9  & 51.2	& 51.9			 \\
\textbf{Fine-Tuned} & 71.4	& 82.7	& 95.8	& 70.4	& 46.5	& 53.0  & 85.3	& 44.4	& 95.0 & 65.3 & 75.1	& 71.7			 \\
\midrule
\textbf{Averaging}\pub{ICML22} & 57.9	& 81.2	 & 58.3	& 43.3	& 39.1	& 40.0 & 80.9	& 40.1	& 92.4  & 52.8  & 53.8	& 55.0			\\
\textbf{Fisher Merging}\pub{NeurIPS22} &  62.2  &  83.3  &  83.3  &  45.9  &  41.0  &  42.2  &  83.1  &  42.2  &  94.1  &  58.3  &  56.7  &  54.2       \\
\textbf{RegMean}\pub{ICLR23} & 58  &  81.2  &  58.3  &  43.3  &  39.2  &  40.2  &  80.9  &  40.1  &  92.5   &  53.5  &  53.8  &  55    \\
\textbf{Task Arithmetic}\pub{ICLR23} & 63.9  &  74.1  &  83.3  &  60.8  &  49.4  &  50.0  &  87.5  &  41.5  &  \textbf{95.3}  &  49.3  &  62.8  &  49.1       \\
\textbf{Ties-Merging}\pub{NeurIPS23}  & 66.8	& 78.6	& \textbf{87.5}	& 66.6	& \textbf{51.3}	& \textbf{51.5}   & 81.7	& 43.2	& 90.9  & 57.6   & 67.0	& 58.4			 \\
\textbf{Consensus Ties}\pub{ICML24}  & 66.6	& 78.5	& 87.3	& 66.4	& 51.1	& 51.2   & 81.6	& \textbf{43.4}	& 90.2  & 57.3   & 67.1	& 58.3			 \\
\textbf{\ourapproach (ours)} & \textbf{68.2}	& 80.1	 & 83.5	 & \textbf{67.3}	& 51.2	& 49.8 & \textbf{88.4}	& 42.6	& 92.8  & 61.9  & \textbf{67.5}	& \textbf{64.8}			 \\

\bottomrule
\end{tabular}
}
\vspace{10pt}
\end{table*}

\begin{table*}[htb!]
\centering
\belowrulesep=0pt
\aboverulesep=0pt
\caption{ Test set performance when merging ViT-B/32 models on 8 vision tasks. Please refer to Section 4.4 for more details.}
\label{tab:app_vit_base}
\resizebox{0.85\linewidth}{!}{  
\begin{tabular}{r|c|cccccccc}

\toprule
\rowcolor{mygray}\textbf{Task($\rightarrow$)}  &   & \multicolumn{8}{c}{\textbf{Test Set Performance}} \\ 
\cline{3-10}
\rowcolor{mygray}\textbf{Method($\downarrow$)} & \multirow{-2}{*}{\textbf{Average}}  & SUN397  &  Cars  &  RESISC45  &  EuroSAT  &  SVHN  &  GTSRB  &  MNIST & DTD \\
\midrule
\textbf{Individual} & 90.5  &  75.3  &  77.7  &  96.1  &  99.7  &  97.5  &  98.7  &  99.7  &  79.4 \\
\textbf{Multitask} & 88.9  &  74.4  &  77.9  &  98.2  &  98.9  &  99.5  &  93.9  &  72.9  &  95.8 \\
\midrule
\textbf{Averaging}\pub{ICML22} & 65.8  &  65.3  &  63.4  &  71.4  &  71.7  &  64.2  &  52.8  &  87.5  &  50.1 \\
\textbf{Fisher Merging}\pub{NeurIPS22}  & 68.3  &  \textbf{68.6}  &  \textbf{69.2}  &  70.7  &  66.4  &  72.9  &  51.1  &  87.9  &  \textbf{59.9} \\
\textbf{RegMean}\pub{ICLR23}  & 71.8  &  65.3  &  63.5  &  75.6  &  78.6  &  78.1  &  67.4  &  93.7  &  52 \\
\textbf{Task Arithmetic}\pub{ICLR23}  & 70.1  &  63.8  &  62.1  &  72  &  77.6  &  74.4  &  65.1  &  94  &  52.2 \\
\textbf{Ties-Merging}\pub{NeurIPS23}  & 73.6  &  64.8  &  62.9  &  74.3  &  78.9  &  83.1  &  71.4  &  97.6  &  56.2 \\
\textbf{Consensus Ties}\pub{NeurIPS23}  & 73.3  &  64.5  &  63.0  &  74.1  &  78.5  &  83.0  &  71.1  &  96.9  &  55.8 \\
\textbf{\ourapproach (ours)}   & \textbf{76.5}  &  66.8  &  65.4  &  \textbf{78.5}  &  \textbf{79.2}  &  \textbf{86.5}  &  \textbf{77.1}  &  \textbf{98.1}  &  59.3 \\

\bottomrule
\end{tabular}
}
\vspace{10pt}
\end{table*}

\begin{table*}[htb!]
\centering
\belowrulesep=0pt
\aboverulesep=0pt
\caption{ Test set performance when merging ViT-L/14 models on 8 vision tasks. Please refer to Section 4.4 for more details.}
\label{tab:app_vit_large}
\resizebox{0.85\linewidth}{!}{  
\begin{tabular}{r|c|cccccccc}
\toprule
\rowcolor{mygray}\textbf{Task($\rightarrow$)}  &   & \multicolumn{8}{c}{\textbf{Test Set Performance}} \\ 
\cline{3-10}
\rowcolor{mygray}\textbf{Method($\downarrow$)} & \multirow{-2}{*}{\textbf{Average}}  & SUN397  &  Cars  &  RESISC45  &  EuroSAT  &  SVHN  &  GTSRB  &  MNIST & DTD \\
\midrule
\textbf{Fine-tuned} & 94.2  &  82.3  &  92.4  &  97.4  &  100  &  98.1  &  99.2  &  99.7  &  84.1  \\
\textbf{Multitask} & 93.5  &  90.6  &  84.4  &  99.2  &  99.1  &  99.6  &  96.3  &  80.8  &  97.6  \\
\midrule
\textbf{Averaging}\pub{ICML22} & 79.6  &  72.1  &  81.6  &  82.6  &  91.9  &  78.2  &  70.7  &  97.1  &  62.8  \\
\textbf{Fisher Merging}\pub{NeurIPS22}  & 82.2  &  69.2  &  \textbf{88.6}  &  87.5  &  93.5  &  80.6  &  74.8  &  93.3  &  70  \\
\textbf{RegMean}\pub{ICLR23}  & 83.7  &  73.3  &  81.8  &  86.1  &  97  &  88  &  84.2  &  98.5  &  60.8  \\
\textbf{Task Arithmetic}\pub{ICLR23} & 84.5  &  74.1  &  82.1  &  86.7  &  93.8  &  87.9  &  86.8  &  98.9  &  65.6  \\
\textbf{Ties-Merging}\pub{NeurIPS23}  & 86  &  76.5  &  85  &  89.4  &  95.9  &  90.3  &  83.3  &  99  &  68.8  \\
\textbf{Consensus Ties}\pub{NeurIPS23}  & 86.2  &  76.6  &  85.2  &  \textbf{89.5}  &  96.3  &  \textbf{90.4}  &  83.6  &  \textbf{99.1}  &  68.8  \\
\textbf{\ourapproach (ours)}  & \textbf{87.6}  &  \textbf{76.8}  &  86.1  &  \textbf{89.5}  &  \textbf{96.5}  &  88.4  &  \textbf{91.1}  &  98.5  &  \textbf{73.7}  \\
\bottomrule
\end{tabular}
}
\vspace{10pt}
\end{table*}

\section{Additional Analysis}
\label{app:b}
\subsection{Ablation Studies}
\label{app_ablation}
Our method incorporates several key factors, including the number of subspaces, the volume of the calibration dataset, and the sparsity of pruning levels. We conducted ablation studies on these elements, with the results presented in Appendix Table~\ref{tab: ablation1},~\ref{tab: ablation2},~\ref{tab: ablation3}. Specifically, we tested our approach on knowledge fusion across eight ViT models for vision tasks.
\begin{table}[ht]
\centering
\caption{The performance of \ourapproach in knowledge fusion on vision tasks across varying volumes of calibration datasets.}
\label{tab: ablation1}
\resizebox{0.75\linewidth}{!}{
\begin{tabular}{cccccc}
\toprule
Volume & Ties-Merging & 1/4 & 1/2 & 1 \\
\midrule
ViT-B/32 & 73.6 & 75.9 & 76.3 & 76.5 \\
\midrule
ViT-L/14 & 86.0 & 87.1 & 87.5 & 87.6 \\
\bottomrule
\end{tabular}
}
\end{table}
\begin{table}[ht]
\centering
\caption{The performance of \ourapproach in knowledge fusion on vision tasks across varying numbers of subspaces.}
\label{tab: ablation2}
\resizebox{0.75\linewidth}{!}{
\begin{tabular}{cccccc}
\toprule
Numbers & Ties-Merging & 1 & 2 & 4 & 8 \\
\midrule
ViT-B/32 & 73.6 & 74.8 & 75.6 & 76.2 & 76.5 \\
\midrule
ViT-L/14 & 86.0 & 86.9 & 87.3 & 87.5 & 87.6 \\
\bottomrule
\end{tabular}
}
\end{table}
\begin{table}[ht]
\centering
\caption{The performance of \ourapproach in knowledge fusion on vision tasks with different sparsity pruning ratios $r$.}
\label{tab: ablation3}
\resizebox{0.75\linewidth}{!}{
\begin{tabular}{cccccc}
\toprule
Ratios & 0.03 & 0.05 & 0.1 & 0.2 & 0.3 \\
\midrule
ViT-B/32 & 75.8 & 76.5 & 76.3 & 75.2 & 72.1 \\
\midrule
ViT-L/14 & 86.9 & 87.6 & 87.2 & 86.5 & 83.4 \\
\bottomrule
\end{tabular}
}
\end{table}
% \paragraph{Number of Subspaces}
% \paragraph{Calibration Dataset}
% \paragraph{Ratios}

\subsection{Hyperparameters}
Due to the hyperparameter sensitivity in task vector-based model merging methods, we provide the optimal values of $\lambda$ and $r$ as determined by our experiments, as outlined in Tab.~\ref{tab:hyperpara_settings}. For Task Arithmetic, we explored $\lambda$ within the range of 0.2 to 1.5, using a step size of 0.1. In the cases of TIES-Merging and \ourapproach, we varied the mask ratios $r$ across \{0.05, 0.1, 0.2\}, while $\lambda$ was searched within the range of 0.8 to 2.5 with a step size of 0.1. For knowledge compression using \ourapproach, we fixed the ratio $r$ at 0.05 to minimize storage costs.
\begin{table*}[ht]
\centering
\vspace{2mm}
\captionsetup{type=table}
\caption{$\lambda$ and pruning ratio $r$ for \ourapproach}
\label{tab:hyperpara_settings}
\resizebox{0.9\linewidth}{!}{  
\begin{tabular}{r|cc|c|c|cc}
\thickhline
\rowcolor{mygray}
Task ($\rightarrow$)   & \multicolumn{2}{c|}{\small7 \textbf{NLP} Tasks} &\small 11 \textbf{PEFT} Tasks & \small3 \textbf{LLM} Tasks & \multicolumn{2}{c}{\small8 \textbf{Vision} Tasks} \\ 
\cline{2-7}
% \cmidrule(lr){7-8}
% \hline
\rowcolor{mygray}
Method ($\downarrow$)  & T5-Base & T5-Large & (IA)$^3$ & LLaMa2 & ViT-B/32 & ViT-L/14\\

% \hline
Task Arithmetic\pub{ICLR23} [$\lambda$] & 0.4 & 0.5  & 0.5 & 0.3 & 0.3 & 0.3 \\ 
Ties-Merging\pub{NeurIPS23} [$\lambda$, $r$] & [1.7, 0.1]  & [2.4, 0.05]  & [1.7, 0.1] & [1.0, 0.1] & [1.0, 0.1] & [1.1, 0.05] \\ 
NPS for fusion (ours) [$\lambda$, $r$] & [1.9, 0.05] & [2.2, 0.05]  & [1.8, 0.1] & [0.9, 0.1] & [1.2, 0.05] & [1.2, 0.05] \\
NPS for compression (ours) [$r$] & 0.05 & 0.05  & - & 0.05 & 0.05 & 0.05 \\
\thickhline
\end{tabular}
}
\end{table*}

% \pub{NeurIPS23} \cite{ties}

\subsection{Time cost}

\begin{table*}[ht]
\centering
\caption{Time Costs for \ourapproach.}
\label{tab:time_cost}
\resizebox{0.9\linewidth}{!}{  
\begin{tabular}{c|ll|l|l|ll}
\thickhline
\rowcolor{mygray}
Task ($\rightarrow$)  & \multicolumn{2}{c|}{\small7 \textbf{NLP} Tasks} &\small 11 \textbf{PEFT} Tasks & \small3 \textbf{LLM} Tasks & \multicolumn{2}{c}{\small8 \textbf{Vision} Tasks} \\ 
% \cline{3-7}
% \cmidrule(lr){7-8}
% \hline
\rowcolor{mygray}
Method ($\downarrow$)  & T5-Base & T5-Large & ~~~~~~~(IA)$^3$ & ~~LLaMa2 & ViT-B/32 & ViT-L/14\\

\hline

Time for Pruning    & 5 secs  & 9 secs  & 1 secs & 113 secs & 4 secs & 7 secs \\
Time for Validation  & 4 mins & 7 mins  & 15 mins & 12 mins & 6 mins & 9 mins \\ 
Generations    & 30 & 50  & 20 & 20 & 30 & 30 \\ 
Total Time for \ourapproach  & 126 mins & 358 mins  & 300 mins & 278 mins & 183 mins & 273 mins \\ 
\thickhline
\end{tabular}
}
\end{table*}
The total time required for the overall \ourapproach strategy is 
\begin{equation}
    T_{\text{total}} = \text{Generations} \times (T_{\text{pruning}} + T_{\text{validate}})
\end{equation} 
where generations represents the number of generations needed for searching, which is a pre-set value and varies with different experiment settings. 
The pruning time mainly depends on the number of model parameters and the size of the model population, while the validation time primarily depends on the volume of inference data and the inference speed. 
We have organized and reported the number of generations and the time required for each task in Appendix Table~\ref{tab:time_cost}. As shown, our method typically requires only a few hours (2-6 hours) to complete, even for large language models.

\section{Implementation details}
\label{app:c}
\subsection{Computational Resources and Runtimes}
\label{app_computation}
Our experiments were conducted on Nvidia A6000 GPUs with 48GB of RAM. 
Depending on the dataset size, fine-tuning the T5-Base and T5-Large models for single tasks took between 15 minutes and 2 hours, while fine-tuning the multitask checkpoint took around eight hours. 
The fine-tuned (IA)$^3$ models were provided by \citet{ties}.\footnote{\url{https://github.com/prateeky2806/ties-merging}}. 
We also used vision models ViT-B/32 and ViT-L/14 as provided by \citet{ilharco2022editing}.\footnote{\url{https://github.com/mlfoundations/task_vectors\#checkpoints}}.
Merge experiments were highly efficient, with evaluations for RoBerta-base, T5-Base, T5-Large, ViT-B/32, and ViT-L/14 models taking less than 2 minutes. However, two specific experiments required more time: (1) Evaluating (IA)$^3$ models took about one hour for 11 datasets due to the need to use multiple templates from prompt sources and compute median results across them. (2) Validation on LLMs (LLaMa2) was also slow, usually requiring about 40 minutes for evaluating 3 datasets.

\subsection{Training details}
\label{app_training_details}
We trained the T5-base and T5-large models for up to 75,000 steps, using a batch size of 1024 and a learning rate of 0.0001. Early stopping with a patience of 5 was employed to prevent overfitting. Training was conducted in bfloat16 to conserve GPU memory, with a sequence length capped at 128 tokens. For the PEFT configuration of the (IA)$^3$ approach on the T0-3B model, the batch size was set to 16 for training and 32 for evaluation, while maintaining a learning rate of 0.0001. The early stopping patience was extended to 10 due to the model's complexity. We didn't use any learning rate scheduler or weight decay during training. For large language models, we used fine-tuned checkpoints from Huggingface\footnote{\url{https://huggingface.co/}}.

In the cross-domain merging experiments, we fine-tuned the RoBERTa-base model with an initial learning rate of 1e-5 and the T5-base model at 1e-4, using the AdamW optimizer. The learning rate was gradually increased during the first 6\% of training steps, then linearly decreased to zero. Both models were trained with a batch size of 16 over 30 epochs for emotion classification, with performance evaluated at the end of each epoch, resuming from the best checkpoint.

\subsection{Evaluation Metrics}
\label{app:metric}
\paragraph{Normalized Accuracy.} We report both normalized and absolute accuracies. Normalization is based on the accuracy of the individual fine-tuned models.
\begin{equation}
    \text{Acc.} =  \frac{1}{N} \sum_{n=1}^{N} \frac{\underset{x\sim\mu_n}{\text{acc}}\left[f_\text{merged}(x)\right]}{\underset{x\sim\mu_n}{\text{acc}}\left[f_\text{fine-tuned}(x)\right]}
    % Acc. >> Normalized Acc.
\end{equation}
\paragraph{H-Score.} To rigorously evaluate our method's ability to mitigate catastrophic forgetting in MLLMs, we use two key metrics: Average Performance and the H-score~\cite{zhumodel}. The H-score, a novel metric, provides a balanced assessment by calculating the harmonic mean between the average performance on original tasks, $\operatorname{Avg}(P_\text{origin})$, and on target tasks, $\operatorname{Avg}(P_\text{target})$. The formula for the H-score is as follows:
\begin{equation}
P_H = \frac{2 \times \operatorname{Avg}(P_\text{origin}) \times \operatorname{Avg}(P_{target})}{\operatorname{Avg}(P_\text{origin}) + \operatorname{Avg}(P_\text{target})}.
\end{equation}
The H-score was introduced to avoid overemphasizing the performance of original tasks, especially as their number grows.
\paragraph{Storage Cost.}
This section show the calculation of the storage cost for each method in Section 4.5 and Appendix A Tab.~\ref{tab:app_compression}. Let $N$ be the number of tasks, $P$ be the number of all parameters, $P'$ be the number of trainable parameters in the model, and $F$ be the number of frozen parameters in the model. Assuming one float parameter takes 32 bits, for each method, their respective storage cost for $T$ tasks is calculated as:
\begin{itemize}
    \item Fine-tuned models: $32(NP'+F)$. $32NP'$ is for storing $T$ trainable parameters and $32F$ is for storing frozen parameters.
    \item Task arithmetic: $32P$; Stores a single model.
    \item Ties-merging: $32P$; Stores a single model.
    \item Consensus Ties: $32P$; Stores a single model.
    \item Zero-shot: $32P$; Stores a single model.
    \item TALL Mask + Ties: $(64+N)P' + 32F$; $64P'+32F $is for storing zeroshot model and multi-task vector, while $NP'$ is for storing T binary masks.
    \item \ourapproach: $32P+(r*32+1)NP'$; $r$ is the sparsity pruning ratio.
    
\end{itemize}
\section{Baseline details} 
\label{app:d}
We provied a detailed baseline description. Our experiments encompass seven comparison methods:
\begin{itemize}[noitemsep,topsep=0pt,parsep=0pt,partopsep=0pt]
    \item \textbf{Individual} means that each task uses an independent fine-tuned model, which has no interference between tasks, but cannot perform multiple tasks simultaneously.
    \item \textbf{Traditional MTL} collects the original training data of all tasks together to train a multi-task model. It can be used as a reference \textit{upper bound} for model merging work.
    % \item \textbf{Weight Averaging} is the simplest method of model merging, which directly averages the parameters of multiple models. It can be used as a \textit{lower bound} for model merging. 
    \item \textbf{Weight Averaging} is the simplest method of model merging, which directly averages the parameters of multiple models using $\theta_m = \sum_{t=1}^{n} \theta_t / n$, calculating the element-wise mean of all individual models. It can be used as a \textit{lower bound} for model merging. \cite{choshen2022fusing,wortsman2022model}. 

    \item \textbf{Fisher Merging}~\citep{matena2022merging} calculates the Fisher information matrix~\citep{fisher1922mathematical} $\hat{F}_t=\mathbb{E}_{x\sim D_t}\mathbb{E}_{y\sim p_{\theta_t}(y|x)} \nabla_{\theta_t} (\log p_{\theta_t}(y|x_t))^2$ to measure the importance of each parameter when merging models for task $t$, where and model merging is performed according to the guidance of this importance.
    % \item \textbf{RegMean}~\citep{jin2023regmean} imposes a constraint when merging models, that is, the $L_2$ distance between the merged model and a single model is required to be as small as possible. 
    \item \textbf{RegMean}~\citep{jin2023regmean} imposes a constraint when merging models, that is, the $L_2$ distance between the merged model's and the individual models' activations. It computes a least-squares solution as $\theta_m = (\sum_{t=1}^n X_t^TX_t)^{-1} \sum_{t=1}^n (X_t^T X_t \theta_t)$, where $X_t$ is the input activation of the corresponding layer.

    \item \textbf{Task Arithmetic}~\citep{ilharco2022editing} first defines the concept of “task vectors” and merges these vectors into a pre-trained model to execute multi-task learning. The model is produced by scaling and adding the task vectors to the initial model as $\theta_m = \theta_\textrm{init} + \lambda * \sum_{t=1}^n \tau_t$.

    \item \textbf{Ties-Merging}~\citep{ties} further solves the task conflict problem in Task Arithmetic~\citep{ilharco2022editing}. It eliminates redundant parameters and resolves symbol conflicts through three steps: Trim, Elect Sign, and Disjoint Merge.
    
    \item \textbf{AdaMerging} automatically learns a merging coefficient for each layer of each task vector in Task Arithmetic~\citep{ilharco2022editing}.

    \item \textbf{LoraHub}~\citep{huang2023lorahub} employs Low-rank Adaptations to dynamically combine task-specific modules for cross-task generalization, and adapts to new tasks by configuring \( \theta' = \sum_{k=1}^{K} w_k \cdot \theta_k \).
    % without extra parameters.

    \item \textbf{DARE}~\citep{yu2023language} sets the majority of delta parameters to zero and rescale the rest by \( \theta' = \theta \cdot (1/(1-p)) \) where \( p \) is the proportion of delta parameters dropped, therefore efficiently reduces parameter redundancy.
\end{itemize}
\section{Datesets details}
% \section{Dataset details}
\label{app:e}
This section provides a detailed dataset description for our experiments.
\paragraph{NLP Tasks.}
Following TIES-Merging~\citep{ties}, we choose seven datasets for merging NLP models: question answering (QASC \cite{khot2020qasc}, WikiQA \cite{yang2015wikiqa}, and QuaRTz \cite{tafjord2019quartz}), paraphrase identification (PAWS \cite{zhang2019paws}), sentence completion (Story Cloze \cite{sharma2018tackling}), and coreference resolution (Winogrande \cite{winogrande} and WSC~\cite{wsc}).

\paragraph{PEFT Models.}
Following TIES-Merging~\citep{ties}, we use eleven datasets including sentence completion (COPA \citep{copa}, H-SWAG \citep{zellers2019hellaswag}, and Story Cloze \citep{sharma2018tackling} datasets), natural language inference (ANLI \citep{nie2019adversarial}, CB \citep{cb}, and RTE \citep{rte}), coreference resolution (WSC \citep{wsc} and Winogrande \citep{winogrande}), and word sense disambiguation (WiC \citep{wic}).

\paragraph{Vision Tasks.} Following Task Arithmetic~\citep{ilharco2022editing}, we study multi-task model merging on eight image classification datasets below.  Stanford Cars \citep{cars} is a car classification dataset consisting of 196 classes of cars. DTD \citep{dtd} is a texture classification dataset comprising 47 classes. EuroSAT \citep{eurosat} comprises 10 classes of geo-referenced satellite images. GTSRB \citep{gtsrb} includes 43 classes of traffic signs. MNIST \citep{lecun1998mnist} features grayscale images of handwritten digits across 10 classes. RESISC45 \citep{cheng2017remote} encompasses 45 classes of remote sensing image scenes. SUN397 \citep{xiao2016sun} consists of 397 classes of scene images. Lastly, SVHN \citep{netzer2011reading} encompasses 10 classes of real-world digital classification images.
\begin{table}[th]
\setlength{\intextsep}{1pt}
\setlength{\columnsep}{2pt}
% \begin{wraptable}{r}{0.4\textwidth}
\caption{\small{Statistics of emotion classification datasets.}}
\label{tab:data_stats_emotion}
\centering
\resizebox{0.9\linewidth}{!}{  
\begin{tabular}{@{}lrrr@{}}
\toprule
                   & Train  & Dev    & Test   \\ \midrule
\textit{In-domain} &        &        &        \\
DialyDialog        & 72,085 & 10,298 & 20,596 \\
CrowdFlower        & 27,818 & 3,974  & 7,948  \\
TEC                & 14,735 & 2,105  & 4,211  \\
Tales-Emotion      & 10,339 & 1,477  & 2,955  \\
ISEAR              & 5,366  & 766    & 1,534  \\ \midrule
% \bottomrule
\end{tabular}
}
% \end{wraptable}
\end{table}
\paragraph{Emotion Classification.}
In order to investigate the performance of the sentiment classification task, following RegMean~\citep{jin2023regmean}, we selected a diverse and challenging set of datasets. Among them, DailyDialogs~\citep{li2017dailydialog}, CrowdFlower, TEC~\citep{mohammad2012emotional}, Tales-Emotion~\citep{alm2005emotions}, and ISEAR~\citep{scherer1994evidence} is utilized to train domain-specific model. For evaluation, we focus exclusively on the fundamental emotions: anger, disgust, fear, joy, sadness, and surprise. A detailed overview of the datasets and statistics is provided in Tab.~\ref{tab:data_stats_emotion}.
\paragraph{LLMs.}
\begin{itemize}[noitemsep,topsep=0pt,parsep=0pt,partopsep=0pt]
    \item CMMLU \cite{li2024cmmlu} is a comprehensive Chinese evaluation benchmark specifically designed to assess language models' knowledge and reasoning abilities in a Chinese context. It covers 67 topics ranging from basic subjects to advanced professional levels.
    \item GSM8K \cite{cobbe2021training} is a collection of 8.5K high-quality, linguistically varied math word problems from grade school, crafted by skilled human authors. The solutions predominantly require executing a series of basic arithmetic operations ($+$, $-$, $\times$, $\div$) to derive the final answer.
    \item HumanEval \cite{chen2021evaluating} is a dataset for evaluating code generation ability, containing 164 manually crafted programming problems covering aspects such as language understanding, reasoning, algorithms, and simple mathematics.
\end{itemize}

% \clearpage %not allowed
% \bibliography{aaai25}
% \input{files/checklist}
\end{document}